\documentclass{article}
\usepackage{iclr2027_conference,times}
\iclrfinalcopy

\usepackage{graphicx}
\usepackage{capt-of}
\usepackage{booktabs}
\usepackage{array}
\usepackage{amsmath}
\usepackage{amssymb}
\usepackage{amsthm}
\usepackage{algorithm}
\usepackage{algpseudocode}
\usepackage{multirow}
\usepackage{placeins}
\usepackage{xcolor}
\usepackage{hyperref}
\usepackage{url}

\definecolor{accent}{RGB}{0,0,180}
\hypersetup{colorlinks=true, linkcolor=accent, citecolor=accent, urlcolor=accent,
            filecolor=accent, pdfborder={0 0 0}}
\hypersetup{pdftitle={Selecting the Minimizer: Test-Time Depth Completion Beyond the Measured Pixels},
            pdfauthor={Yuhyun Kim, TaeSeop Kim, Dong-Geol Choi}}

\newtheorem{proposition}{Proposition}[section]

\newcommand{\R}{\mathbb{R}}
\newcommand{\Msk}{\mathcal{M}}
\newcommand{\Var}{\mathrm{Var}}
\newcommand{\Cov}{\mathrm{Cov}}

\newcommand{\loss}{\mathcal{L}}
\newcommand{\pred}{f_{\theta}}
\newcommand{\Fnorm}[1]{\lVert #1 \rVert_{F}}
\newcommand{\norm}[1]{\lVert #1 \rVert}
\newcommand{\proj}{\mathcal{P}}
\newcommand{\res}{\tilde{q}}
\newcommand{\Lap}{\mathbf{L}}
\newcommand{\Deg}{\mathbf{D}}
\newcommand{\Aff}{\mathbf{W}}

\newcommand{\sws}{SWS}
\newcommand{\srp}{SRP}

\newcommand{\lora}{LoRA-TTO}
\newcommand{\testprompt}{TestPromptDC}

\newenvironment{tightlist}
  {\begin{itemize}\setlength{\itemsep}{2pt}\setlength{\parskip}{0pt}%
   \setlength{\topsep}{3pt}\setlength{\parsep}{0pt}}
  {\end{itemize}}

\newcommand{\fittable}[1]{%
  \resizebox{\ifdim\width>\linewidth \linewidth\else \width\fi}{!}{#1}}

\newcommand{\best}[1]{\underline{\textbf{#1}}}

\title{Selecting the Minimizer: Test-Time Depth\\Completion Beyond the Measured Pixels}

\author{Yuhyun Kim, TaeSeop Kim \& Dong-Geol Choi \\
Department of Information and Communication Engineering \\
Hanbat National University, Daejeon, Republic of Korea \\
\texttt{\char`\{yuhyun.kim,taeseop.kim\char`\}@edu.hanbat.ac.kr}, \texttt{dgchoi@hanbat.ac.kr}}

\begin{document}

\maketitle
\lhead{Preprint}

\vspace{-1.0em}
\begin{center}
\includegraphics[width=\linewidth]{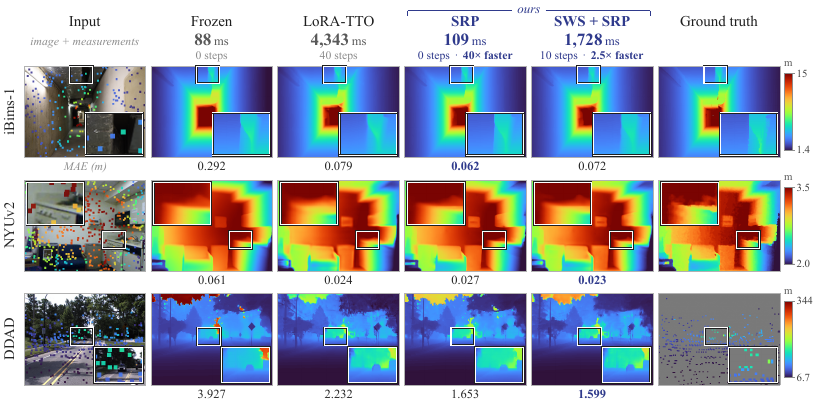}
\end{center}
\vspace{-0.4em}
\captionof{figure}{\textbf{Depth completion at three budgets: the reconstructions look alike,
the cost does not.} Each head gives the latency at VGA on one RTX 4090 and the adaptation
steps (speed-ups against \lora{}), and each panel that sample's MAE (m), the lowest in bold. Rows are iBims-1, NYUv2 and DDAD, each \textbf{the first image of
its benchmark}; the white box marks the window magnified in the corner. Depth scale is fixed
per row (meters, right), and the input measurements are dilated for visibility. Accuracy over
all 6{,}479 images is in Tab.~\ref{tab:main}.}
\label{fig:teaser}
\vspace{0.2em}

\begin{abstract}
Recent test-time optimization for depth completion adapts a frozen monocular metric-depth
model to the sparse measurements of a single image through a low-rank adapter, and so needs
no completion training set. Eliminating its scale-and-shift alignment analytically puts the
objective in closed form, which explains why the procedure is fragile and slow: its minimizers
leave at least $95\%$ of the parameter dimensions free, so the difficulty is not
reaching a minimizer but choosing one, and its standard zero initialization is degenerate, with
one factor's gradient and curvature vanishing identically. We add one component for each.
Semiparametric residual propagation (SRP) chooses the minimizer explicitly, in function space,
splitting the correction into a parametric term in the predicted depth and a field harmonic on
a bandwidth-free affinity graph, with the measurements imposed as constraints. A spectral warm
start (SWS) replaces the degenerate initialization with the best rank-$r$ descent step, whose
amplitude is computed rather than searched. Across five benchmarks, $10$ adaptation steps---a
quarter of the baseline's budget---match or beat the full $40$-step baseline on every metric
while lowering MAE by $27\%$ on KITTI and $31\%$ on DDAD. \srp{} alone needs
no backward pass, runs $\mbox{30--40}\times$ faster than that adaptation and lowers its
MAE by $22\%$ on KITTI and $26\%$ on DDAD. Code is available at \url{https://github.com/yuhyun00/selecting-the-minimizer-test-time-depth-completion}.
\end{abstract}

\section{Introduction}
\label{sec:intro}

Depth completion is normally supervised, and the resulting models inherit the sensor pattern,
depth range and scene statistics of their training set. A recent alternative needs no
completion training set: a frozen monocular metric-depth
model~\citep{piccinelli2025unidepthv2,wang2025moge2,lin2025depthanything3} already predicts a
plausible dense surface, so at test time one fits a low-rank adapter~\citep{hu2022lora} to a
\emph{single} image until the prediction agrees with that image's sparse measurements, and
reads out the adapted dense prediction~\citep{seo2026efficient}. Nothing is learned across
images, so the same procedure applies to a lidar scan, a structured-light frame and a handful
of visual-inertial points.

Low-rank test-time optimization (\lora{}) is a cheap route to depth
completion---$\mbox{1.6--4.4}\times$ faster than prompt tuning~\citep{jeong2025testprompt}
across the resolutions we time (Tab.~\ref{tab:cost}), and $5.1\times$ faster than guided
diffusion~\citep{viola2025marigolddc} at VGA (Tab.~\ref{tab:cost})---yet every
image still costs tens of parameter updates, and the result depends on the seed. We argue that
speed and stability are one problem, and that both follow from the structure of the test-time
objective.

The structure is visible once the objective is written in closed form. Every test-time
objective of this kind aligns the prediction to the measurements by scale and shift before
taking the residual; that inner problem has an exact solution, and substituting it back
(Sec.~\ref{sec:analysis}) reduces the objective to the variance of the measurements times one
minus their squared correlation with the prediction. Three consequences follow. Every
gradient carries the units of the data, so no step length stated in absolute terms can suit
more than one benchmark. The standard zero initialization of the adapter is degenerate: the
objective is constant in one factor there, so every run spends its first steps leaving that
point in a direction chosen by the seed. And because the measurements cover at most a few
percent of the pixels, the minimizers form a set retaining at least $95\%$ of the
parameter dimensions: the optimizer does not fail to find a minimizer, and \emph{which} one it
reaches---hence what the unmeasured pixels look like---is left to implicit bias.

The last two point to fixes, and by the first, neither fix carries a constant in data units.
Semiparametric residual propagation (\srp{}, Sec.~\ref{sec:srp}) makes the choice of minimizer
explicit, in function space, where it is unique: it writes the correction to the aligned
prediction as a parametric term in the predicted depth plus a field harmonic on an
image-and-depth affinity graph, and imposes the measurements as constraints rather than
penalties; the affinity is built from gradient ranks, so it has no bandwidth to tune. The
spectral warm start (\sws{}, Sec.~\ref{sec:sws}) replaces the degenerate initialization with
the best rank-$r$ descent step at the frozen weights, with its direction, per-layer scale and
amplitude all computed in closed form. Because \srp{} is a read-out, it applies to whichever
prediction reaches it---frozen, or adapted at any budget---so the two components trace one
accuracy/cost curve (Fig.~\ref{fig:teaser}) rather than a single operating point.

Across five benchmarks and 6{,}479 images, no training-free method run under our protocol lies below that
curve on either metric (Sec.~\ref{sec:cost}). With no backward pass at all, \srp{} on the
frozen prediction is $\mbox{30--40}\times$ faster end to end than the $40$-step
adaptation and lowers its MAE on both driving benchmarks; with $10$ steps, a quarter of the
baseline's budget, \sws{}~$+$~\srp{} matches or beats it on every benchmark and metric, lowering
MAE by $27\%$ on KITTI and $31\%$ on DDAD. The analysis also predicts
what should \emph{not} help---better optimization and parameter-space regularization---and
none of twelve such interventions improves on the baseline by more than half a percent
(Sec.~\ref{sec:parts}).

\paragraph{Contributions.}
\textbf{(i)}~The closed form of the test-time objective and three exact consequences: a
gradient scale set by the data, a degenerate standard initialization and a minimizer set of
codimension at most $4.3\%$ (Sec.~\ref{sec:analysis}; proofs in
App.~\ref{sec:appx-proofs}). \textbf{(ii)}~\srp{}, which chooses among those minimizers in
function space, with the measurements as constraints, an affinity with no length scale and a
parametric term that escapes the maximum principle bounding any propagated correction
(Sec.~\ref{sec:srp}). \textbf{(iii)}~\sws{}, an analytic first step whose direction,
per-layer scale and amplitude are all computed rather than searched (Sec.~\ref{sec:sws}).

\section{Related Work}
\label{sec:related}

\paragraph{Depth completion without a completion training set.}
Recent monocular estimators predict metric depth across domains without per-dataset
fine-tuning~\citep{piccinelli2024unidepth,piccinelli2025unidepthv2,yang2024depthanythingv2};
their output is geometrically consistent, but its absolute scale is unreliable---exactly the
gap a handful of measurements can close. Three lines exploit this without paired sparse and
dense depth: test-time prompt tuning~\citep{jeong2025testprompt}, diffusion
priors~\citep{viola2025marigolddc,lee2025zeroshotdc} and low-rank decoder
adaptation~\citep{seo2026efficient}; concurrent work~\citep{ke2026capa} adapts a frozen 3D
foundation model per sample in the same way. We study the last, the cheapest of the three at VGA and
the one whose objective admits an exact characterization; the analysis applies to any method
whose per-sample objective is an aligned masked regression against the measurements.

\paragraph{Adapter initialization.}
Among works that change how low-rank factors~\citep{hu2022lora} are initialized or
scaled~\citep{meng2024pissa,hayou2024loraplus,liu2024dora}, closest to \sws{} is
LoRA-GA~\citep{wang2024loraga}, which also initializes from the singular vectors of the
gradient at the frozen weights, as GaLore~\citep{zhao2024galore} does for its projection
subspace. Three differences are forced by our analysis: LoRA-GA subtracts its initial product
so that the starting function is unchanged, whereas \sws{} takes the step; \sws{} scales each
layer by its own Frobenius norm rather than by one absolute length; and its amplitude is
computed, not tuned. Curvature spectra~\citep{li2018visualizing} and sharpness-aware
minimization~\citep{foret2021sam} treat flat directions as a proxy for generalization over a
dataset; here they are exact and known in closed form, and flatness-seeking does not help
(Sec.~\ref{sec:parts}).

\paragraph{Propagation and harmonic interpolation.}
Propagating sparse values along image affinities is
classical~\citep{levin2004colorization,he2010guidedfilter}, as is edge-aware fitting under a
data penalty in a bilateral grid~\citep{barron2016fbs}; we do not claim the propagation itself.
Supervised depth completion
learns the affinities from paired data~\citep{cheng2018cspn,park2020nlspn}, which this setting
rules out. The closest formulation comes from semi-supervised learning: \citet{zhu2003harmonic}
pose label propagation as a harmonic function on a weighted graph with the labeled points as
Dirichlet boundary values. \srp{} departs from it in three ways. \textbf{(a)}~Their affinity
carries a length scale per feature, learned by entropy minimization; our rank affinity \eqref{eq:rank} has none.
\textbf{(b)}~Their field \emph{is} the estimate, while ours corrects one---semiparametric, as in
regression kriging~\citep{hengl2007regressionkriging}, but with a trend in the predicted depth
rather than in position, so it is defined far from every measurement. \textbf{(c)}~The maximum
principle is a proof device there; here it is a design constraint whose effect we measure and
whose limit the parametric term escapes. The same ceiling binds penalized edge-aware
solvers: the fast bilateral solver's output is a weighted average of its targets, so applied to
the residual it cannot leave the measured range either. Tuned per benchmark on ground
truth, it reaches $1.03$ of the $40$-step baseline's MAE against \srp{}'s $0.93$ on the frozen
prediction (App.~\ref{sec:appx-fbs}).

\section{What the Test-Time Objective Actually Is}
\label{sec:analysis}

Every statement in this section is exact: App.~\ref{sec:appx-proofs} proves each one, and
App.~\ref{sec:appx-verify} verifies them numerically and measures the size of the effects they
force. Sec.~\ref{sec:closed} derives the closed form and its three consequences,
(C1)--(C3); Sec.~\ref{sec:manifold} shows that choosing a minimizer, not finding one, is the
hard part. Notation is collected in Tab.~\ref{tab:notation} (App.~\ref{sec:appx-proofs}).

\paragraph{Setup.}
\label{sec:setup}
Let $\Msk$ be the set of $m$ pixels of an image carrying a depth measurement, $y \in \R^{m}$
those measurements, and $\pred \in \R^{N}$ the dense prediction of the frozen model, whose
decoder carries a low-rank adapter with parameters $\theta \in \R^{D}$ adding
$\Delta W = s\,AB$ (scale $s$, rank $r$) at each adapted layer (App.~\ref{sec:appx-setup}). Because
monocular depth is defined only up to an affine map for many backbones, the objective aligns
the prediction to the measurements before taking the residual:
\begin{equation}
  \loss(\theta) \;=\; \min_{a,b}\ \frac{1}{m}\sum_{i \in \Msk}
     \bigl(a\,[\pred]_i + b - y_i\bigr)^{2}.
  \label{eq:obj}
\end{equation}
Implementations recompute $(a,b)$ at every step from the detached prediction and treat it as
constant during backpropagation. This is not an approximation: $(a,b)$ is the exact inner
minimizer, so by the envelope theorem the detached gradient is the exact gradient of $\loss$
(Prop.~\ref{prop:envelope}).

\subsection{The objective is a correlation criterion}
\label{sec:closed}

Solving the inner least-squares problem and substituting back gives
(Prop.~\ref{prop:closed})
\begin{equation}
  \loss(\theta) \;=\; \Var_{\Msk}(y)\,
    \Bigl(1 - \rho_{\Msk}\bigl(\pred, y\bigr)^{2}\Bigr),
  \label{eq:closed}
\end{equation}
whenever $\Var_\Msk(\pred) > 0$. The identity is the classical coefficient of determination;
our observation is that it \emph{is} the test-time objective---the alignment is the standard
scale-and-shift-invariant depth criterion~\citep{ranftl2022midas}---so minimizing the aligned
squared error is exactly maximizing the correlation between the prediction and the
measurements on $\Msk$. Three consequences follow: (C1) and (C3) are the first two named in
Sec.~\ref{sec:intro}, and (C2) determines how curvature must be measured.

\paragraph{(C1) The gradient carries the units of the data.}
$\Var_\Msk(y)$ multiplies the whole objective, hence every gradient and every curvature
eigenvalue. Across the benchmarks the loss at initialization spans nearly six orders of
magnitude, while the dimensionless factor $1-\rho_\Msk^2$ spans about two: most of the
apparent difficulty gap between an indoor room and a highway is a change of units, and a step
rule in absolute units cannot suit more than one benchmark. Adaptive optimizers mask this
without removing it, spending steps to estimate from gradient magnitudes a factor that
\eqref{eq:closed} gives in closed form.

\paragraph{(C2) The objective has two exact flat manifolds.}
Since \eqref{eq:closed} depends on $\pred$ only through a correlation,
$\loss(\alpha \pred + \beta) = \loss(\pred)$ for any $\alpha \neq 0$ and $\beta$; separately, the
low-rank product is invariant under $(A,B) \mapsto (AM, M^{-1}B)$ for \emph{any} invertible
$r \times r$ matrix $M$, an $r^2$-dimensional flat manifold per adapted layer
(Prop.~\ref{prop:flat}). Motion along the affine manifold is wasted, since the alignment
reabsorbs it, and curvature must be computed with the alignment differentiated, or the
apparent sharpest direction is mostly a change of output scale (App.~\ref{sec:appx-verify}).

\paragraph{(C3) The standard initialization is blind in one factor and indefinite in
curvature.}
Low-rank adapters set $A = 0$ so that the adapted model starts at the frozen model. Since
$\Delta W = s\,AB \equiv 0$ for \emph{any} $B$ when $A = 0$, the objective is constant in $B$
there, not merely stationary: $\nabla_{\!B}\loss \equiv 0$ and $\nabla^2_{\!BB}\loss \equiv 0$.
All seeds therefore start at the same point in function space and differ only in a direction
the objective cannot see. Nor is the point a minimum: $\nabla_{\!A}\loss =
s\,(\nabla_{\!\Delta W}\loss)B^{\!\top}$ does not vanish---it is the gradient \sws{} builds its
step from---and with the $BB$ block identically zero, the Hessian is indefinite whenever the
frozen model's gradient is nonzero (Prop.~\ref{prop:saddle}). A short run must grow $A$ out of
this point and still converge within the same budget.

\subsection{The hard part is choosing a minimizer, not finding one}
\label{sec:manifold}

It is tempting to read the instability above as a hard nonconvex problem in
$D = 438{,}784$ dimensions; the measurements say otherwise. The minimizer set
\begin{equation}
  \Theta^{\star} \;=\; \bigl\{\theta \;:\; \pred|_{\Msk} = \alpha y + \beta \mathbf{1},
    \ \alpha \neq 0\bigr\}
  \label{eq:manifold}
\end{equation}
imposes $m-2$ conditions on $D$ parameters whenever $y$ is nonconstant on $\Msk$; wherever
$\theta \mapsto \pred|_\Msk$ is a submersion---which we check rather than assume
(App.~\ref{sec:appx-verify})---it is a submanifold of codimension exactly $m-2$
(Prop.~\ref{prop:codim}). The hypothesis enters only this dimension count; the closed form,
(C1)--(C3), \sws{} and \srp{} do not use it. Where it fails, the Jacobian's kernel only grows,
so to first order at least $D - m$ directions remain free and the count is conservative. Over the evaluated images $m$ runs from $82$ to $18{,}966$, so the
codimension lies between $0.02\%$ and $4.3\%$ of $D$: $\Theta^\star$ retains at
least $95\%$ of the parameter dimensions on \emph{every} evaluated image, even the
densest KITTI frame. Adaptation drives $1-\rho_\Msk^{2}$ from a median of $10^{-1}$ to
$2\!\times\!10^{-3}$, yet dense accuracy keeps improving after the objective has effectively
saturated (App.~\ref{sec:appx-verify}): the run drifts along $\Theta^{\star}$, and that drift
changes the unmeasured pixels while leaving the measured ones fitted.

This is the implicit-regularization regime of under-determined factorized
regression~\citep{gunasekar2017implicit}, with a generalization gap that is spatial rather
than statistical. It makes a falsifiable prediction, confirmed in Sec.~\ref{sec:parts}:
interventions that improve the \emph{optimization} or regularize in \emph{parameter} space
should not help, since no parameter-space penalty reduces the dimension of $\Theta^\star$.
Shrinking $\Theta^\star$ requires information about the unmeasured pixels, and without ground
truth there are two sources, the image and the frozen model's own prediction; \srp{} uses both,
and only the second is defined far from every measurement.

\section{Method}
\label{sec:method}

\begin{figure}[t]
\centering
\includegraphics[width=0.92\linewidth]{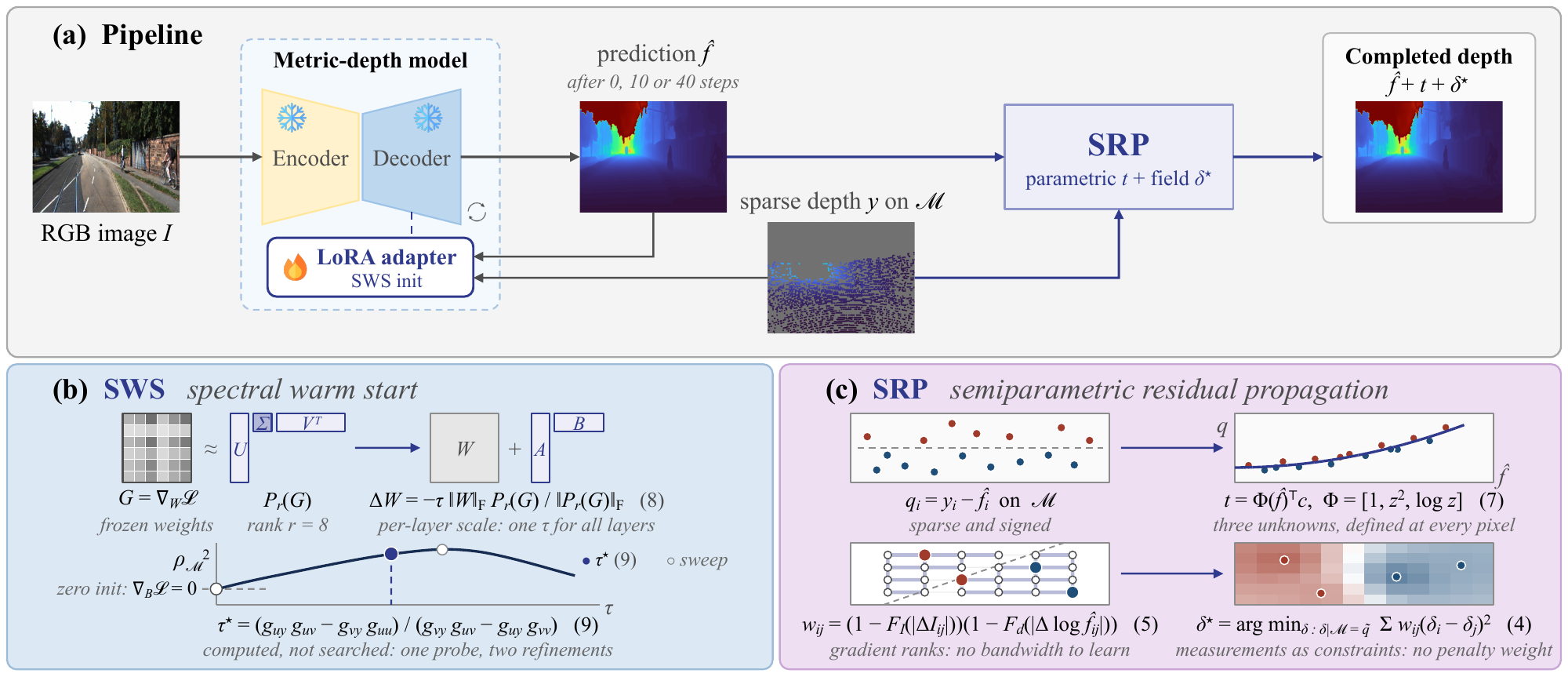}
\caption{\textbf{Where \sws{} and \srp{} sit in one adaptation.} \textbf{(a)}~Pipeline: a frozen monocular metric-depth model, a low-rank adapter
initialized by \sws{}, and \srp{} as the read-out of whichever prediction reaches it; the loop
runs $10$ steps at the point we lead with, $40$ at the baseline's budget and $0$ when skipped.
Images: the KITTI sample of Fig.~\ref{fig:qualitative}, its sparse input and a $10$-step run
before and after the solve. \textbf{(b)}~\sws{}: the step \eqref{eq:sws} and its amplitude
\eqref{eq:tau}, against a sweep of $\tau$. \textbf{(c)}~\srp{}: the residual on $\Msk$, the
parametric term \eqref{eq:trend}, the affinity \eqref{eq:rank} and the constrained field
\eqref{eq:harm}. Dashed outlines: frozen or inherited from \citet{seo2026efficient}; color:
ours.}
\label{fig:overview}
\end{figure}

(C1) constrains our own design as much as the baseline's: any constant in data units---a
penalty weight, an affinity bandwidth, a step length---can suit one benchmark only, so neither
component below carries one. Fig.~\ref{fig:overview} shows where the two sit in one
adaptation: \sws{} replaces the initialization, the adaptation loop is unchanged and \srp{}
is a read-out applied to whichever prediction reaches it.

\subsection{Semiparametric residual propagation}
\label{sec:srp}

Let $\hat{f} = a\pred + b$ be the aligned prediction and $q_i = y_i - \hat{f}_i$ the residual
at the measured pixels. The objective determines $\hat f$ on $\Msk$ and says nothing
elsewhere, so we choose what happens elsewhere explicitly.

\paragraph{A measurement is a constraint, not a preference.}
The obvious formulation trades a data term against a smoothness penalty,
$\sum_{i\in\Msk}(\delta_i-q_i)^2 + \mu\sum_{(i,j)\in\mathcal{N}}
w_{ij}(\delta_i-\delta_j)^2$, with $\mathcal{N}$ the four-neighborhood and $w_{ij}$ an
affinity that is large across a flat edge and small across a sharp one. Unlike the
parameter-space problem, whose solution set has dimension close to $D$, this function-space
problem has exactly one solution---the reason for moving the choice here. But it carries three
constants in data units ($\mu$, the bandwidth $w_{ij}$ needs, and the solver's iteration
count), which (C1) says will not transfer, and $\mu$ has no defensible value at all: a
measurement is a value the field must take, not a term to trade against a prior. Imposing it
as such replaces the data term by a constraint,
\begin{equation}
  \delta^{\star} \;=\; \arg\min_{\delta \,:\, \delta|_{\Msk} = \res}\;
    \sum_{(i,j) \in \mathcal{N}} w_{ij}\,(\delta_i - \delta_j)^{2},
  \label{eq:harm}
\end{equation}
whose solution is graph-harmonic away from $\Msk$: the Dirichlet problem on the affinity graph
with the measurements as boundary values ($\res$ rather than $q$: one more term is subtracted
below). \textbf{This is the same prior with its weight taken to the limit}---letting
$\mu \to 0$ recovers \eqref{eq:harm} exactly (Prop.~\ref{prop:harm})---and a tuned $\mu$ costs $18\%$ of MAE on the frozen
prediction.

\paragraph{An affinity with no bandwidth.}
The usual $w_{ij} = \exp(-\lVert I_i - I_j \rVert / \sigma)$ has the same defect one level
down: $\sigma$ is a threshold on image gradients, whose units the benchmark sets. We use each
gradient's \emph{rank} instead: with $F_I$ and $F_d$ the empirical distributions, over all
edges of one image, of the image gradient and of the predicted log-depth gradient,
\begin{equation}
  w_{ij} \;=\; \bigl(1 - F_{I}(|\Delta I_{ij}|)\bigr)\bigl(1 - F_{d}(|\Delta \log \hat f_{ij}|)\bigr),
  \label{eq:rank}
\end{equation}
so the flattest edge in the image has weight $1$ and the sharpest $0$, and an edge is
crossable only if the image and the predicted geometry both say it is flat. Ranks are
invariant to any monotone rescaling of the gradient---precisely what $\sigma$ was there to
absorb---so the construction has no constant at all, and a tuned bandwidth costs $12\%$ of
MAE.

\paragraph{The maximum principle is a ceiling, and it binds.}
A field harmonic on a graph is a convex combination of its boundary values: with $h_{ij}$ the
probability that a random walk started at $i$ first meets $\Msk$ at $j$
(Prop.~\ref{prop:harm}),
\begin{equation}
  \delta_i \;=\; \sum_{j \in \Msk} h_{ij}\, \res_j,
  \qquad h_{ij} \ge 0, \quad \textstyle\sum_{j} h_{ij} = 1,
  \label{eq:maxprin}
\end{equation}
so $\min_\Msk \res \le \delta_i \le \max_\Msk \res$ at every unmeasured pixel. The fact is
classical~\citep{levin2004colorization,zhu2003harmonic}; we read it as a \emph{ceiling} on
what any propagated correction can do. \textbf{The solve can never correct by more than the
measurements themselves show}, and as $i$ moves away from $\Msk$ the distribution $h_{i\cdot}$
spreads, so the correction flattens. Grouping pixels by distance to the nearest measurement
shows where the ceiling binds (Tab.~\ref{tab:reach}): indoors the solve stays within
$0.97$--$1.31\times$ of adaptation's squared error in every group, while DDAD is ahead
within $4$ pixels and $5.7\times$ worse between $16$ and $64$, and KITTI
shows the same shape at less than half the size. The ceiling binds where the residual is both
large and far from any measurement: the failure is one of reach, not of the prior.

\paragraph{A term the field cannot represent.}
What the field cannot supply is a correction defined far from every measurement, so we fit
one on the measured pixels as a function not of position but of the predicted depth, which is
available at every pixel:
\begin{equation}
  t \;=\; \Phi(\hat f)^{\!\top} c,
  \quad
  \Phi(z) = [\,1,\; z^{2},\; \log z\,],
  \quad
  c = \arg\min_{c}\ \sum_{i \in \Msk}\bigl(\Phi(\hat f_i)^{\!\top} c - q_i\bigr)^{2}.
  \label{eq:trend}
\end{equation}
The affine alignment leaves $q$ orthogonal on $\Msk$ to $\mathrm{span}\{1, \hat f\}$, so a fit
on that span alone returns zero; $\hat f^{2}$ and $\log \hat f$ are the lowest-order terms not
already exhausted, and the two the physics suggests: a monocular backbone's absolute error
grows superlinearly with range, while its relative error is roughly constant. \textbf{This
generalizes scale-and-shift alignment}, which assumes that the residual left by two degrees of
freedom is unstructured; richer bases reach further but generalize worse, and the balance tips
between three terms and four (App.~\ref{sec:appx-srp}).

The solve then propagates only what the parametric term leaves: $\res_i = q_i - t_i$ on $\Msk$
in \eqref{eq:harm}, and the output is $\hat f + t + \delta^\star$. The constraint is
untouched---at $i \in \Msk$ the output is $\hat f_i + t_i + (q_i - t_i) = y_i$---and the two
pieces are complementary: one parametric, global and extrapolating, the other nonparametric,
local and interpolating. What the term buys is reach: on DDAD it cuts the $16$--$64$ pixel
group from $5.7\times$ to $2.3\times$ and improves the nearer groups as
well. \textbf{That it is a function of depth and not of position is what buys this reach}: the same three
unknowns fitted on pixel coordinates---the regression-kriging design---leave that group almost
untouched (App.~\ref{sec:appx-srp}), because a trend on the grid says nothing about a far pixel
that the solve does not. Indoors the term is nearly neutral, and on VOID, whose measured set is
the most biased sample of its scene, it changes MAE by $+1.2\%$.

\paragraph{The iteration count is a budget, not a constant.}
In the penalized form the iteration count is a regularizer that must be tuned
(App.~\ref{sec:appx-alt}); for \eqref{eq:harm}, accuracy is monotone in the work spent. We use a
\emph{cascadic} cycle, every level solving the true problem at its own resolution: a fixed
$4\times 120$ budget comes within $1.1\%$ of the converged field,
and its loop length depends on the resolution alone, so the solve is a static
computation graph (Sec.~\ref{sec:cost}).

\subsection{Spectral warm start}
\label{sec:sws}

\sws{} replaces the zero initialization with an analytic first step: \emph{spectral} because
its direction is a truncated singular value decomposition of the gradient, a \emph{warm start}
because it optimizes a surrogate rather than guessing a convenient size. Zero initialization
suits fine-tuning on a corpus; with one image, a budget of tens of steps and (C3), it is a
degenerate start. LoRA-GA also leaves that point but subtracts its initial product, so its run
still starts at the frozen function; \sws{} takes the step.

\paragraph{Which step, in closed form.}
Let $G_l = \nabla_{W_l}\loss$ be the gradient with respect to the \emph{base} weight of an
adapted layer at the frozen model. Among rank-$r$ matrices of a given Frobenius norm, the one
that decreases the loss fastest to first order maximizes $\langle -G_l, \Delta W\rangle$, and
by von Neumann's trace inequality it is (Prop.~\ref{prop:sws})
\begin{equation}
  \Delta W_l \;=\; -\,\tau\,\Fnorm{W_l}\,
     \frac{\proj_r(G_l)}{\Fnorm{\proj_r(G_l)}},
  \label{eq:sws}
\end{equation}
with $\proj_r$ truncating to the leading $r$ singular triplets. The factor $\Fnorm{W_l}$,
absent from LoRA-GA, cannot be absorbed by any computed amplitude: $\tau$ is one scalar shared
by \emph{every} adapted layer, so per-layer scaling changes the direction of the joint step
(App.~\ref{sec:appx-sws} reports what dropping it costs).

\paragraph{How far, in closed form.}
The perturbation is linear in $\tau$, so to first order the prediction on $\Msk$ along the ray
is $u + \tau v$, with $u = \hat f|_\Msk$ and $v$ the directional derivative, and
\eqref{eq:closed} makes the objective a function of $\rho_\Msk(u + \tau v, y)^2$ alone. With
$\bar{\cdot}$ centering over $\Msk$ and $g_{pq} = \langle \bar p, \bar q\rangle$ the centered
inner product of any two of $u$, $v$ and $y$, that correlation is a ratio of quadratics in
$\tau$ whose unique maximizer is (Prop.~\ref{prop:tau})
\begin{equation}
  \tau^{\star} \;=\;
    \frac{g_{uy}\,g_{uv} - g_{vy}\,g_{uu}}{g_{vy}\,g_{uv} - g_{uy}\,g_{vv}}.
  \label{eq:tau}
\end{equation}
One extra forward pass supplies $v$ by finite difference, and the probe size cancels to first
order. A tangent taken at the origin undershoots when the ray curves ($0.84$ of the
true maximizer indoors and $0.55$ on DDAD, over 60 images), so we re-solve the
same model from where it lands, a secant through the last two points evaluated. Two
refinements beat none, one and three, for $4$ forward passes in total; the count is fixed once
for all five benchmarks (Tab.~\ref{tab:decisions}), and it beats every greedy
amplitude search we tried (App.~\ref{sec:appx-sws}). The adapter realizes \eqref{eq:sws} as
$sAB$ from the truncated singular vectors; both factors are then free and the shipped recipe
runs unchanged, so a warm-started run differs from the reference only in its initialization.

\section{Experiments}
\label{sec:exp}

\subsection{Protocol}
\label{sec:protocol}

Five benchmarks differ in scene type, depth range and---most importantly here---how the sparse
measurements are produced: \textbf{NYUv2}~\citep{silberman2012nyu} and \textbf{iBims-1}
\citep{koch2018ibims} (indoor, structured light and laser scan), \textbf{KITTI}
\citep{uhrig2017sparsity} and \textbf{DDAD} \citep{guizilini2020packnet} (outdoor driving,
lidar), and \textbf{VOID} \citep{wong2020void} (visual-inertial odometry points). Sparse inputs
follow the protocol of \citet{seo2026efficient}; the measured set runs from $82$ to $18{,}966$
pixels, a factor of $231$. We evaluate every image of every split,
6{,}479 in total, after removing the 25 images that an earlier version of the
amplitude rule used for calibration (App.~\ref{sec:appx-setup}). The splits are
unequal---DDAD alone is $69\%$ of the four benchmarks other than VOID, which we call
the core---so the headline is the unweighted mean of per-benchmark means, and we name the
pooled per-image statistic wherever a paired test needs it.

All rows come from one pipeline on one set of inputs: \testprompt{}~\citep{jeong2025testprompt}
is re-implemented and run under our protocol, Marigold-DC~\citep{viola2025marigolddc} is run
with its official code and settings (App.~\ref{sec:appx-setup}), and \citet{lee2025zeroshotdc}
is not run. The frozen backbone is
UniDepthV2 with a ViT-L encoder~\citep{piccinelli2025unidepthv2}, with rank-8 adapters in the
decoder's linear layers and the configuration published with \lora{}: $40$ steps of
AdamW~\citep{loshchilov2019adamw} at a learning rate of $10^{-2}$. Our runs use the same
optimizer, learning rate and, at the full budget, step count, differing only in the
initialization and the read-out. The forward pass is bitwise reproducible and backward
reductions are deterministic; rows that start from \lora{}'s initialization are means over five
seeds, and \sws{} is deterministic.

\textbf{Design selection.} Nothing is tuned per benchmark or per image. The discrete choices
in Tab.~\ref{tab:decisions} were compared on development subsets drawn from these splits. On the
4{,}907 images outside all of them, the $10$-step configuration still matches or beats the
$40$-step baseline in every cell. Holding out any one benchmark leaves the constraint, the
affinity, the warm start and the $10$-step budget unchanged, and every remaining choice stays
within $0.6$ points of its best alternative (App.~\ref{sec:appx-heldout}).

\begin{table}[!t]
\centering
\caption{\textbf{Five benchmarks, one protocol.} MAE and RMSE in meters; lower is better.
``Steps'' is the number of adaptation steps per image. \best{Bold and underlined} marks the
best cell of the lower block; bold row labels mark \srp{} alone and the one we lead with. Rows
that start from \lora{}'s initialization are means over five seeds; the others are
deterministic. All rows come from one pipeline on one set of inputs (6{,}479 images, full
splits), so none is comparable to a published figure. Marigold-DC uses its own diffusion
prior with its official settings (one sample at a fixed seed, $^\dagger$50 denoising steps), on
the same inputs and scored on its own output; its 10-sample ensemble on a fixed subset is in
App.~\ref{sec:appx-more}.}
\label{tab:main}
\footnotesize
\fittable{\begin{tabular}{lccccccccccc}
\toprule
 & & \multicolumn{2}{c}{iBims-1} & \multicolumn{2}{c}{VOID} & \multicolumn{2}{c}{NYUv2} & \multicolumn{2}{c}{KITTI} & \multicolumn{2}{c}{DDAD} \\
\cmidrule(lr){3-4}\cmidrule(lr){5-6}\cmidrule(lr){7-8}\cmidrule(lr){9-10}\cmidrule(lr){11-12}
Method & Steps & MAE & RMSE & MAE & RMSE & MAE & RMSE & MAE & RMSE & MAE & RMSE \\
\midrule
\multicolumn{12}{l}{\emph{Depth foundation models}} \\
UniDepthV2~\citep{piccinelli2025unidepthv2} & 0 & 0.302 & 0.395 & 0.238 & 0.594 & 0.306 & 0.429 & 1.539 & 3.560 & 5.176 & 14.843 \\
\quad + scale\,\&\,shift & 0 & 0.100 & 0.216 & 0.184 & 0.555 & 0.145 & 0.253 & 1.405 & 2.977 & 5.774 & 13.866 \\
Depth Anything V2~\citep{yang2024depthanythingv2} & 0 & 0.390 & 0.494 & 0.334 & 0.680 & 1.165 & 1.263 & 2.301 & 3.986 & 7.334 & 17.580 \\
\quad + scale\,\&\,shift & 0 & 0.137 & 0.258 & 0.202 & 0.568 & 0.205 & 0.311 & 1.705 & 3.313 & 7.438 & 13.903 \\
Depth Anything 3~\citep{lin2025depthanything3} & 0 & 1.112 & 1.222 & 0.737 & 1.049 & 0.979 & 1.081 & 2.200 & 3.772 & 16.004 & 24.504 \\
\quad + scale\,\&\,shift & 0 & 0.110 & 0.222 & 0.190 & 0.555 & 0.158 & 0.264 & 1.462 & 3.041 & 4.631 & 10.476 \\
\midrule
\multicolumn{12}{l}{\emph{Zero-shot depth completion, diffusion prior}} \\
Marigold-DC~\citep{viola2025marigolddc}, 1 sample & 50$^\dagger$ & 0.063 & 0.203 & 0.156 & 0.551 & 0.059 & 0.150 & 0.576 & 1.683 & 2.967 & 7.872 \\
\midrule
\multicolumn{12}{l}{\emph{Zero-shot depth completion, UniDepthV2 backbone}} \\
Frozen decoder $+$ scale\,\&\,shift & 0 & 0.100 & 0.216 & 0.191 & 0.574 & 0.173 & 0.280 & 1.419 & 3.017 & 5.888 & 14.088 \\
$+$ \testprompt{}~\citep{jeong2025testprompt} & --- & 0.053 & 0.165 & 0.170 & 0.534 & 0.134 & 0.236 & 1.123 & 2.367 & 5.973 & 13.212 \\
$+$ \lora{}~\citep{seo2026efficient} & 40 & 0.041 & 0.146 & 0.136 & 0.505 & 0.054 & 0.125 & 0.438 & 1.372 & 2.002 & 5.937 \\
$+$ \sws{} \emph{(ours)} & 40 & 0.040 & 0.147 & 0.133 & 0.502 & 0.052 & 0.124 & 0.399 & 1.357 & 1.752 & 5.846 \\
\midrule
$+$ \textbf{\srp{}} \emph{(ours)} & \textbf{0} & 0.040 & 0.162 & 0.152 & 0.539 & 0.055 & 0.142 & 0.340 & 1.448 & 1.480 & 6.325 \\
$+$ \textbf{\sws{} $+$ \srp{}} \emph{(ours)} & \textbf{10} & 0.041 & \best{0.144} & 0.134 & 0.504 & 0.054 & 0.125 & 0.321 & \best{1.301} & 1.386 & \best{5.494} \\
$+$ \lora{} $+$ \srp{} \emph{(ours)} & 40 & \best{0.037} & 0.146 & 0.132 & 0.503 & 0.050 & 0.123 & 0.318 & 1.325 & 1.399 & 5.613 \\
$+$ \sws{} $+$ \srp{} \emph{(ours)} & 40 & \best{0.037} & 0.146 & \best{0.130} & \best{0.501} & \best{0.049} & \best{0.122} & \best{0.310} & 1.329 & \best{1.341} & 5.695 \\
\bottomrule
\end{tabular}
}
\end{table}

\subsection{Main result}
\label{sec:main}

Tab.~\ref{tab:main}'s upper block reports three metric-depth backbones raw and after a
scale-and-shift fit to the same measurements; everything below is measured on top of that
alignment. \textbf{The lower block is a cost curve, not a ranking}: rows differ in what they
spend as well as in what they achieve. Relative differences are signed, so negative is an
improvement. These claims concern training-free completion; supervised methods need paired
training data (Sec.~\ref{sec:related}). One sample of the diffusion prior
Marigold-DC~\citep{viola2025marigolddc} trails $40$-step \lora{} on every benchmark and metric
at $5.1\times$ its latency; its 10-sample ensemble, at ten times that cost, stays $7\%$ behind
on MAE on a fixed subset and leads only on NYUv2 (Tab.~\ref{tab:main}, Tab.~\ref{tab:cost},
App.~\ref{sec:appx-more}).

\textbf{At equal budget, the read-out improves every benchmark on both metrics.} Adding \srp{}
to the shipped $40$-step run improves both metrics on all five benchmarks, and MAE on
$100.0\%$ of the core's images. The mean of per-benchmark relative MAE is
$-18.5\%$ over the core and $-15.3\%$ over all five; per image, which is
what a paired bootstrap applies to, it is $-26.5\%$ ($[-26.7,\,-26.2]$, $p\,<10^{-4}$). The
two differ because DDAD carries both the most images and the largest gain
(App.~\ref{sec:appx-results}). The smallest gains are on RMSE for VOID and iBims-1, below half a
percent (the table rounds the latter to a tie): a squared metric prices the tail, which the
solve reaches least.

\begin{figure}[t]
\centering
\includegraphics[width=\linewidth]{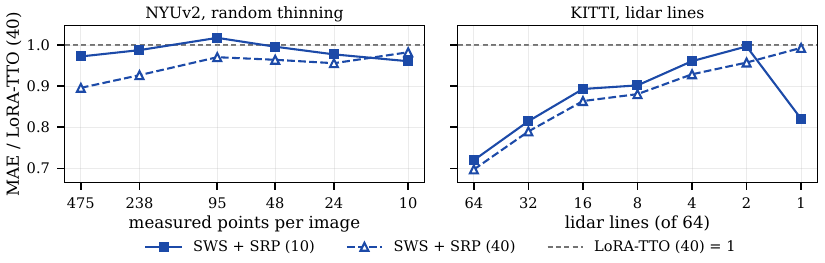}
\caption{\textbf{Accuracy as measurements thin out.} MAE relative to the $40$-step
\lora{} on the same images and the same thinned measurements (100 held-out images per
benchmark; below 1 is better). Left: random thinning of NYUv2's points. Right: KITTI's 64-line
lidar thinned to fewer lines. \srp{} alone and \lora{} $+$ \srp{} are in Tab.~\ref{tab:density}.}
\label{fig:density}
\end{figure}

\textbf{At a quarter of the budget, it matches or beats the baseline everywhere.} \sws{} with
$10$ steps, read out through \srp{}, is the configuration we lead with: at least as accurate as
the full $40$-step \lora{} on all five benchmarks and both metrics, at a quarter of the
steps and $-15.1\%$ MAE over the core. Indoors it ties or wins
narrowly---no cell moves by as much as $2\%$, and on $4$ of the $10$ cells the
paired per-image difference is not distinguishable from zero ($p \geq 0.08$); on the two
driving benchmarks it lowers MAE by $27\%$ and $31\%$ and RMSE by
$5\%$ and $7\%$, with every remaining cell at $p \leq 0.0038$.
\textbf{Where the gain falls is predicted}: KITTI and DDAD have the largest scenes and the
largest remaining errors, where the choice among minimizers decides most
(Sec.~\ref{sec:manifold}) and the maximum principle binds (Tab.~\ref{tab:reach}). Nor is it a
density effect alone: its Spearman correlation is $+0.34$ with a sample's measurement density
but $-0.50$ with the baseline's error on it (App.~\ref{sec:appx-results}).
Thinning the measurements leaves this ordering in place (Fig.~\ref{fig:density},
App.~\ref{sec:appx-more}): on NYUv2 the $10$-step configuration falls behind at a single
density of the sweep, 95 measured pixels (by $1.7\%$), and leads again below it; on KITTI it leads at every
density and down to a single scan line ($-18\%$ MAE).

\textbf{With no adaptation at all, the read-out is a different operating point.} Applied to the
frozen prediction, \srp{} needs no backward pass, adapter or optimizer, and is ahead of \lora{}
on MAE on three of the five benchmarks. It is behind on RMSE on all five, because it leaves a
tail it cannot reach and a squared metric is priced by that tail.

\textbf{The components transfer across backbones.} Because Sec.~\ref{sec:analysis} concerns
the objective rather than the backbone, App.~\ref{sec:appx-results} repeats the leading rows on
MoGe-2 with nothing retuned: the $40$-step \sws{}~$+$~\srp{} result carries to both backbones and wins every
driving cell on each, and at $10$ steps the driving gains persist while a few indoor cells move
the wrong way by at most $+1.1\%$. \textbf{The accuracy/cost curve
transfers; where a given backbone sits on it does not.}

\begin{figure}[t]
\centering
\includegraphics[width=\linewidth]{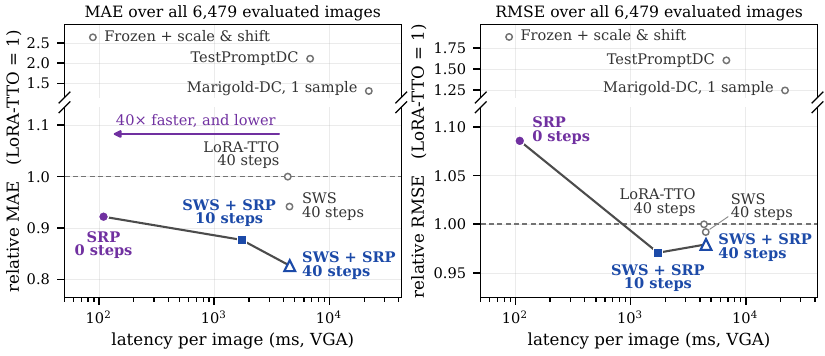}
\caption{\textbf{Tab.~\ref{tab:main} and Tab.~\ref{tab:cost} on one pair of axes.} Each
point is a configuration's error relative to \lora{} (mean over the five benchmarks, all
6{,}479 images) against its latency at VGA on one RTX 4090; the dashed line is the
$40$-step baseline, and lower-left is better. \textbf{The $y$ axis is broken} between the
adapted and unadapted configurations. The Marigold-DC point is one sample at its default
768-pixel processing resolution (Tab.~\ref{tab:cost}).}
\label{fig:costcurve}
\end{figure}

\subsection{Cost}
\label{sec:cost}

Every method is timed by us on one NVIDIA RTX 4090 under one protocol (Tab.~\ref{tab:cost}).
\textbf{The adaptation was the cost, and it no longer is.} Everything \srp{} adds---affinity,
fit and captured solve---totals 21~ms at VGA, against 88~ms for the unavoidable
forward pass and 4{,}343~ms for a full $40$-step adaptation, and stays at 5--24\% of
the forward pass across resolutions. The remaining cost is therefore the backbone's:
\srp{} on the frozen prediction is $\mbox{30--40}\times$ faster end to end than the
adaptation, and the $10$-step configuration $\mbox{2.5--2.7}\times$ faster at equal or
better accuracy. Because \eqref{eq:harm}'s loop length is fixed by the resolution alone, we
capture the solve in a CUDA graph and replay it, reproducing the uncaptured result bit for bit
at $\mbox{5--10}\times$ lower cost (App.~\ref{sec:appx-results}). One capture serves every
input of a given shape, so a fixed-resolution stream captures once; a new shape costs a
one-time $0.8$--$0.9$~s to warm up and capture.

\textbf{Accuracy and cost on one pair of axes.} In Fig.~\ref{fig:costcurve} the three points
we ship form a \emph{frontier}---no training-free method run under our protocol lies below it on either
metric---whose shape differs between metrics: on MAE it falls monotonically with budget from
the adaptation-free point, while on RMSE the adaptation-free point sits \emph{above} the
baseline and the $10$-step point is the first below it. \textbf{That crossing is why we report
a budget rather than an operating point}: which point to deploy depends on the latency
available and on how heavily the tail is priced, while the baseline's single point is dominated
by the $10$-step configuration.

\subsection{What each part is worth, and what does not work}
\label{sec:parts}

\textbf{Each component earns its place at a different end of the curve.} Removing one design
decision from \srp{} at a time (App.~\ref{sec:appx-srp}): constraints, ranked affinities and
image structure are worth $18\%$, $12\%$ and $27\%$ on the frozen
prediction, and all three are within $1.6\%$ of full \srp{} after $40$ steps: they
carry information a long way, and after adaptation the residual is small and local. The
parametric term is the sharpest case: it is worth $13\%$ of RMSE
on DDAD with no adaptation, and after $40$ steps it changes no benchmark's MAE in the third
decimal place. \sws{} is the mirror image (App.~\ref{sec:appx-sws}): aligning the
initialization with the gradient's leading singular subspace, as LoRA-GA does, is worth
$3.9\%$ at $40$ steps and $8.6\%$ at $10$---the steps a
zero-initialized run spends leaving the degenerate point of (C3) are a fixed cost---and taking
the step adds $3.4\%$ and $5.3\%$. Every \sws{} variant has a seed
standard deviation of exactly zero, against $0.049$ for the baseline at $10$ steps.

\textbf{The analysis makes a falsifiable prediction, and it holds.} Sec.~\ref{sec:manifold}
predicts that interventions improving the optimization, or regularizing in parameter space,
should not help. We ran twelve, each applied to \lora{} with everything else
fixed---schedules, factor-asymmetric learning rates, sharpness-aware minimization, averaging,
early stopping, proximal and trust-region terms among them. Ten come in at or above parity;
the only two below it move along the exactly flat directions of (C2), and even they are within
half a percent, as expected for changes the objective cannot see. Two further
negatives sharpen the method. Restricting the adapter to the leading singular subspace of $G_l$
reaches $1.15\times$ the baseline's MAE, so \sws{} frees both factors after the spectral
step. And no training loss folds the solve into the adapter: asked only to \emph{copy} the
converged field, the adapter reaches $1.737$ against the field's $1.123$ (relative to
\sws{}~$+$~\srp{}), so \textbf{the minimizer that \eqref{eq:harm} selects lies outside what a
rank-8 adapter can represent}, and the two stages stay separate (App.~\ref{sec:appx-alt}).

\section{Conclusion}
\label{sec:conclusion}

The test-time objective of low-rank depth completion is exactly a correlation criterion scaled
by the variance of the measurements. Its gradients therefore carry the units of the data, its
standard initialization is degenerate and its minimizers retain at least $95\%$ of the
parameter dimensions on every image, so the difficulty is choosing a minimizer, not reaching
one. \sws{} computes the first step, and \srp{} chooses among the minimizers in function space,
with the measurements as constraints, an affinity with no length scale and a parametric
term in the predicted depth. Together they yield
an accuracy/cost curve---a read-out with no backward pass, a $10$-step configuration that matches
or beats the $40$-step baseline on every benchmark and metric, and an equal-budget configuration
that improves on it everywhere---that transfers across frozen backbones. The analysis was tested
rather than asserted: optimization and parameter-space interventions do not help, as it
predicts, and the maximum principle binds exactly where residuals are large and far from any
measurement.

\textbf{Limitations.}
\label{sec:limitations}
Without adaptation, \srp{} still trails the $40$-step run $4$ to $16$ pixels from a
measurement (Tab.~\ref{tab:reach}), and VOID's tracker-placed points are a biased sample of the
scene (App.~\ref{sec:appx-limits}). The parametric basis is the lowest-order extension of the
affine alignment, refitted on every image; richer bases trade reach for generalization
(App.~\ref{sec:appx-srp}). Affinities robust to such feature-driven sparsity and an
extrapolating field that converges at this budget are natural next steps, both about how to
choose the unmeasured pixels---the question the closed form identifies as the one that matters.

\section*{AI Use Statement}

We used AI tools to assist with language editing, including grammar correction
and refinement of the academic writing style, as well as ablation experiment
design and ablation code implementation. The core research ideas, methodology,
and main experiments were developed and conducted by the authors. All
AI-assisted outputs were reviewed and tested by the authors, who take full
responsibility for the final manuscript and code.

\section*{Reproducibility Statement}

\textbf{Analysis.} Every statement of Sec.~\ref{sec:analysis} and Sec.~\ref{sec:method} is proved
in App.~\ref{sec:appx-proofs}, and App.~\ref{sec:appx-verify} checks on every sample of a
diagnostic subset that the implemented objective is the one analyzed, including the envelope
identity, both flat manifolds of (C2) over the full gauge group, the vanishing curvature block
of (C3) and the submersion hypothesis behind Sec.~\ref{sec:manifold}. \textbf{Method.}
App.~\ref{sec:appx-method} gives the linear system of \eqref{eq:harm}, the cascadic hierarchy,
the parametric fit and pseudocode for \sws{} (Alg.~\ref{alg:sws}); Tab.~\ref{tab:hparams} lists
the full configuration, which has no calibration set and no constant in data units, and whose
discrete choices are listed with their development subsets in Tab.~\ref{tab:decisions}
(App.~\ref{sec:appx-setup}). The captured and uncaptured solves agree
bitwise, the forward pass is bitwise reproducible and backward reductions are deterministic.
\textbf{Evaluation.} App.~\ref{sec:appx-setup} states the datasets, the per-benchmark image
counts, the removal of the 25 calibration images, the metrics and statistics and the
timing protocol. Tab.~\ref{tab:main} and Tab.~\ref{tab:full} evaluate every image of every split (6{,}479
images), all rows come from one pipeline on one set of inputs and every latency is measured by us;
the other studies run on stratified subsets of the splits. Negative results are
reported with their sample sizes and sweep extents in App.~\ref{sec:appx-alt}. \textbf{Code.}
Source code for the method, the benchmark harness and every table and figure generator is
available at \url{https://github.com/yuhyun00/selecting-the-minimizer-test-time-depth-completion}, with a provenance file mapping each table and figure to
the script and result files that produce it; three result sets (the foundation-model and
\testprompt{} rows of Tab.~\ref{tab:main}, the \testprompt{} row of Tab.~\ref{tab:cost} and
Tab.~\ref{tab:folding}) regenerate from shipped intermediates rather than raw inputs.

\newpage
\appendix
\section*{Appendix}

The appendix is organized as follows.
\begin{tightlist}
\item \textbf{App.~\ref{sec:appx-proofs}} proves the statements of Sec.~\ref{sec:analysis} and
  Sec.~\ref{sec:method}.
\item \textbf{App.~\ref{sec:appx-method}} gives the implementation of \srp{} and \sws{}, with
  pseudocode.
\item \textbf{App.~\ref{sec:appx-setup}} describes datasets, splits, metrics, hyperparameters,
  baselines and the timing protocol.
\item \textbf{App.~\ref{sec:appx-results}} extends the main results: all four metrics, paired
  statistics, a second backbone, the budget sweep and the latency breakdown.
\item \textbf{App.~\ref{sec:appx-verify}} verifies the analysis numerically and measures the size
  of its three consequences.
\item \textbf{App.~\ref{sec:appx-srp}} and \textbf{App.~\ref{sec:appx-sws}} ablate \srp{} and
  \sws{}, one design decision at a time.
\item \textbf{App.~\ref{sec:appx-alt}} reports design alternatives that do not help, first those
  the analysis predicts should fail and then others a reader may consider.
\item \textbf{App.~\ref{sec:appx-qual}} shows qualitative results, and \textbf{App.~\ref{sec:appx-limits}}
  discusses limitations.
\item \textbf{App.~\ref{sec:appx-more}} adds three comparisons: Marigold-DC on a fixed subset,
  sensitivity to measurement density, structured missingness and placement, and the fast bilateral solver
  on the residual.
\end{tightlist}
Diagnostic studies run on stratified subsets of the evaluation splits. They are read down a column rather than against Tab.~\ref{tab:main}, whose
rows are the full splits.

\FloatBarrier
\section{Derivations}
\label{sec:appx-proofs}

\begin{table}[!htbp]
\centering
\caption{\textbf{Notation.} Symbols of Sec.~\ref{sec:analysis} and Sec.~\ref{sec:method},
in order of appearance.}
\label{tab:notation}
\footnotesize
\begin{tabular}{@{}ll@{}}
\toprule
Symbol & Meaning \\
\midrule
$\Msk$, $m$ & the measured pixels of one image, and their number \\
$y \in \R^{m}$ & the measurements on $\Msk$ \\
$\pred \in \R^{N}$ & dense prediction of the frozen model with adapter parameters $\theta \in \R^{D}$ \\
$\Delta W = s\,AB$ & adapter update of one layer, rank $r$ and scale $s$ \\
$\loss(\theta)$ & aligned test-time objective \eqref{eq:obj} \\
$\Var_\Msk$, $\rho_\Msk$ & variance and correlation over $\Msk$ \\
$\Theta^{\star}$ & the minimizer set \eqref{eq:manifold} \\
$\hat f = a\pred + b$ & the aligned prediction \\
$q$, $\res$ & residual on $\Msk$, before and after the parametric term \\
$w_{ij}$ & rank affinity of edge $(i,j)$ \eqref{eq:rank} \\
$\delta^{\star}$ & constrained field \eqref{eq:harm} \\
$h_{ij}$ & probability that a walk from $i$ first meets $\Msk$ at $j$ \eqref{eq:maxprin} \\
$t = \Phi(\hat f)^{\!\top}c$ & parametric term \eqref{eq:trend} \\
$G_l$, $W_l$ & gradient and base weight of adapted layer $l$; $\proj_r$ keeps $r$ singular triplets \\
$\tau$, $\tau^{\star}$ & amplitude of the warm start \eqref{eq:sws} and its closed form \eqref{eq:tau} \\
$u$, $v$ & prediction on $\Msk$ and its directional derivative along the step \\
\bottomrule
\end{tabular}
\end{table}

This section proves the statements of Sec.~\ref{sec:analysis} and Sec.~\ref{sec:method}. For
$p \in \R^{m}$ indexed by $\Msk$, let $\bar p = p - \tfrac{1}{m}(\mathbf{1}^{\!\top}p)\mathbf{1}$ be
its centered version, $\Var_\Msk(p) = \tfrac{1}{m}\norm{\bar p}^{2}$,
$\Cov_\Msk(p,q) = \tfrac{1}{m}\langle \bar p, \bar q\rangle$ and
$\rho_\Msk(p,q) = \Cov_\Msk(p,q)/\sqrt{\Var_\Msk(p)\Var_\Msk(q)}$. For fixed $\theta$ we write
$f = \pred|_\Msk$.

\begin{proposition}[Closed form, Eq.~\eqref{eq:closed}]
\label{prop:closed}
If $\Var_\Msk(f) > 0$, the inner problem of \eqref{eq:obj} has the unique minimizer
$a^\star = \Cov_\Msk(f,y)/\Var_\Msk(f)$, $b^\star = \tfrac1m\mathbf 1^{\!\top}(y - a^\star f)$, and
$\loss(\theta) = \Var_\Msk(y)\,\bigl(1-\rho_\Msk(f,y)^{2}\bigr)$.
\end{proposition}
\begin{proof}
For fixed $a$ the optimal $b$ is the mean of $y - af$, which leaves
$\tfrac1m\norm{a\bar f - \bar y}^2 = a^2\Var_\Msk(f) - 2a\Cov_\Msk(f,y) + \Var_\Msk(y)$. This is
strictly convex in $a$ when $\Var_\Msk(f)>0$ and is minimized at $a^\star$, with value
$\Var_\Msk(y) - \Cov_\Msk(f,y)^2/\Var_\Msk(f) = \Var_\Msk(y)\,(1-\rho_\Msk^2)$.
\end{proof}

\begin{proposition}[The detached gradient is exact]
\label{prop:envelope}
Let $\ell(\theta,a,b) = \tfrac1m\sum_{i\in\Msk}(a[\pred]_i + b - y_i)^2$ with $\pred$
differentiable in $\theta$. On $\{\theta : \Var_\Msk(\pred) > 0\}$,
$\nabla\loss(\theta) = \nabla_{\theta}\,\ell(\theta,a,b)\big|_{(a,b) = (a^\star(\theta),\,b^\star(\theta))}$.
\end{proposition}
\begin{proof}
By Prop.~\ref{prop:closed}, $a^\star$ and $b^\star$ are differentiable functions of $\theta$
on this set. The chain rule gives
$\nabla\loss = \nabla_\theta\ell + \partial_a\ell\,\nabla a^\star + \partial_b\ell\,\nabla b^\star$,
and $\partial_a\ell = \partial_b\ell = 0$ at the inner minimizer.
\end{proof}

\begin{proposition}[Invariances, (C2)]
\label{prop:flat}
(i)~$\loss$ is unchanged when $\pred$ is replaced by $\alpha\pred + \beta\mathbf 1$ for any
$\alpha\neq0$ and $\beta$. (ii)~For an adapted layer with $\Delta W = sAB$,
$A\in\R^{d_{\mathrm{out}}\times r}$ and $B\in\R^{r\times d_{\mathrm{in}}}$, the action
$(A,B)\mapsto(AM, M^{-1}B)$ of $M \in \mathrm{GL}(r)$ leaves $\Delta W$, hence $\loss$,
unchanged, and the orbit of any $(A,B)$ with $\operatorname{rank}A = r$ is an
$r^2$-dimensional submanifold.
\end{proposition}
\begin{proof}
(i)~Centering removes $\beta$, and $\alpha$ multiplies $\Cov_\Msk$ by $\alpha$ and
$\sqrt{\Var_\Msk}$ by $|\alpha|$, so $\rho_\Msk^2$ is unchanged. (ii)~$AMM^{-1}B = AB$. If
$\operatorname{rank}A = r$, then $AM = A$ forces $M = I$, so the stabilizer is trivial and the
orbit has the dimension of $\mathrm{GL}(r)$.
\end{proof}

\begin{proposition}[The zero initialization, (C3)]
\label{prop:saddle}
Let $G$ be the gradient of $\loss$ with respect to the base weight of an adapted layer at the
frozen model, and consider any point with $A = 0$. Then (i)~$\loss$ is constant in $B$ on
$\{A=0\}$, so $\nabla_{\!B}\loss = 0$ and $\nabla^2_{\!BB}\loss = 0$ there; (ii)~$\nabla_{\!A}\loss =
s\,G B^{\!\top}$; and (iii)~if $G \neq 0$, the Hessian of $\loss$ with respect to $(A,B)$ is
indefinite.
\end{proposition}
\begin{proof}
(i)~$\Delta W = sAB = 0$ for every $B$ when $A = 0$. (ii)~This is the chain rule through
$\Delta W = sAB$, with $\nabla_{\Delta W}\loss = G$ at $\Delta W = 0$. (iii)~Perturb $(0,B)$ to
$(\varepsilon a, B + \varepsilon b)$. Then $\Delta W = s\varepsilon\,aB + s\varepsilon^2 ab$, and
the second-order term of $\loss$ is $\varepsilon^2 Q(a,b)$ with
$Q(a,b) = s\langle G, ab\rangle + \tfrac{s^2}{2}\,\mathcal H[aB, aB]$, where $\mathcal H$ is the
Hessian of $\loss$ in $W$. Choose $a = u e_1^{\!\top}$ and $b = e_1 v^{\!\top}$ with
$u^{\!\top}Gv \neq 0$, so that $\langle G, ab\rangle \neq 0$. As $t \to 0$,
$Q(ta, \pm b) = \pm\, t\, s\langle G,ab\rangle + O(t^2)$ takes both signs.
\end{proof}

\begin{proposition}[Dimension of the minimizer set]
\label{prop:codim}
Let $F:\R^D\to\R^m$, $\theta\mapsto\pred|_\Msk$, be smooth, let $y$ be nonconstant on $\Msk$, and
let $\Theta^\star = F^{-1}(S)$ with $S = \{\alpha y + \beta\mathbf 1 : \alpha\neq0,\ \beta\in\R\}$.
Then $\Theta^\star$ is exactly the zero set of $\loss$, and if it is nonempty and $F$ is a
submersion at each of its points, it is an embedded submanifold of dimension $D - m + 2$.
\end{proposition}
\begin{proof}
$\loss = 0$ if and only if $\rho_\Msk^2 = 1$, i.e., $\bar f \parallel \bar y$, i.e., $f\in S$
($\alpha\neq0$ because $\Var_\Msk(f)>0$). Since $y$ and $\mathbf 1$ are linearly independent, $S$
is an embedded two-dimensional submanifold of $\R^m$. A submersion is transverse to every
submanifold, so $F^{-1}(S)$ is an embedded submanifold of codimension $m - 2$.
\end{proof}
With $D = 438{,}784$ and $m \in [82, 18{,}966]$, the codimension $(m-2)/D$ lies between
$0.02\%$ and $4.3\%$, which is the statement of Sec.~\ref{sec:manifold};
App.~\ref{sec:appx-verify} tests the submersion hypothesis.

\begin{proposition}[Steepest rank-$r$ step, Eq.~\eqref{eq:sws}]
\label{prop:sws}
For $G \neq 0$ and $c > 0$,
$\max\{\langle -G, X\rangle : \operatorname{rank}X\le r,\ \Fnorm{X} = c\} = c\,\Fnorm{\proj_r(G)}$,
attained at $X = -c\,\proj_r(G)/\Fnorm{\proj_r(G)}$.
\end{proposition}
\begin{proof}
By von Neumann's trace inequality and the Cauchy--Schwarz inequality over the at most $r$
nonzero singular values of $X$,
$\langle -G, X\rangle \le \sum_{i\le r}\sigma_i(G)\,\sigma_i(X)
\le \bigl(\sum_{i\le r}\sigma_i(G)^2\bigr)^{1/2}\Fnorm{X} = c\,\Fnorm{\proj_r(G)}$, and
$\langle G, \proj_r(G)\rangle = \Fnorm{\proj_r(G)}^2$ shows that the bound is attained.
\end{proof}
Eq.~\eqref{eq:sws} is this maximizer with $c = \tau\Fnorm{W_l}$.

\begin{proposition}[The amplitude, Eq.~\eqref{eq:tau}]
\label{prop:tau}
Let $u, v, y\in\R^m$ with $\bar u$ and $\bar v$ linearly independent, $g_{pq} = \langle\bar p,\bar q\rangle$,
and $R(\tau) = \rho_\Msk(u + \tau v, y)^2$. If $\Delta = g_{vy}g_{uv} - g_{uy}g_{vv} \neq 0$, then
$\tau^\star = (g_{uy}g_{uv} - g_{vy}g_{uu})/\Delta$ is the unique global maximizer of $R$ on $\R$.
If $\Delta = 0$, $R$ has no maximizer on $\R$.
\end{proposition}
\begin{proof}
$R(\tau)$ is the squared cosine between $\bar u + \tau\bar v$ and $\bar y$. Over the plane
$\Pi = \mathrm{span}\{\bar u,\bar v\}$, the squared cosine with $\bar y$ is maximized exactly on
the nonzero multiples of the projection $p = \alpha\bar u + \beta\bar v$ of $\bar y$ onto $\Pi$,
where Cramer's rule, with $\det = g_{uu}g_{vv} - g_{uv}^2 > 0$, gives
$\alpha\det = g_{vv}g_{uy} - g_{uv}g_{vy} = -\Delta$ and $\beta\det = g_{uu}g_{vy} - g_{uv}g_{uy}$.
Every vector of $\Pi$ with nonzero $\bar u$-coefficient is a multiple of $\bar u + \tau\bar v$ for
exactly one $\tau$, and squared cosines are invariant to scaling. If $\Delta\neq0$, then
$\alpha\neq0$, so $p$ is reached at $\tau^\star = \beta/\alpha$, which is the stated ratio, and at
no other $\tau$. If $\Delta = 0$, then $p \parallel \bar v$ is approached only as $|\tau|\to\infty$.
\end{proof}
The case $\Delta = 0$ is the guard of App.~\ref{sec:appx-method}.

\begin{proposition}[The constrained solve, Eqs.~\eqref{eq:harm} and~\eqref{eq:maxprin}]
\label{prop:harm}
Let $\Lap = \Deg - \Aff$ be the Laplacian of the affinity graph, $U$ the unmeasured pixels and
$P = \Deg^{-1}\Aff$, and suppose every pixel of $U$ has a positive-weight path to $\Msk$. Then
(i)~$\Lap_{UU}$ is symmetric positive definite and \eqref{eq:harm} has the unique solution
$\delta_U = \Lap_{UU}^{-1}\Aff_{U\Msk}\,\res$; (ii)~$\delta_U = H\res$ with
$H = (I - P_{UU})^{-1}P_{U\Msk}$ entrywise nonnegative and $H\mathbf 1 = \mathbf 1$, where $H_{ij}$, the $h_{ij}$ of \eqref{eq:maxprin},
is the probability that the random walk with transition matrix $P$ started at $i$ first
enters $\Msk$ at $j$; and (iii)~the minimizer of the penalized objective
$\sum_{i\in\Msk}(\delta_i - \res_i)^2 + \mu\,\delta^{\!\top}\Lap\delta$ converges to $\delta^\star$ as
$\mu\to0$.
\end{proposition}
\begin{proof}
(i)~The stationarity condition of \eqref{eq:harm} in $\delta_U$ is
$\Lap_{UU}\delta_U = -\Lap_{U\Msk}\res = \Aff_{U\Msk}\res$. $\Lap_{UU}$ is symmetric and weakly
diagonally dominant with a nonpositive off-diagonal, and every connected component of $U$
contains a pixel adjacent to $\Msk$, whose row is strictly dominant; hence $\Lap_{UU}$ is
nonsingular, and, being symmetric with a nonnegative spectrum by Gershgorin's theorem,
positive definite. (ii)~$\Lap_{UU} = \Deg_{UU}(I - P_{UU})$, and $P_{UU}$ is sub-stochastic with
spectral radius below one under the connectivity assumption, so
$(I-P_{UU})^{-1} = \sum_{k\ge0} P_{UU}^k \ge 0$; this series is the first-passage decomposition of
the walk, and $P_{UU}\mathbf 1 + P_{U\Msk}\mathbf 1 = \mathbf 1$ gives $H\mathbf 1 = \mathbf 1$.
(iii)~The penalized minimizer solves $(\Lap + \mu^{-1}\Pi_\Msk)\,\delta = \mu^{-1}\Pi_\Msk\res$, with
$\Pi_\Msk$ the coordinate projection onto $\Msk$ and $\res$ extended by zero; as $\mu\to0$ its
restriction to $\Msk$ tends to $\res$ and its restriction to $U$ tends to the solution of (i).
\end{proof}

\paragraph{Orthogonality of the aligned residual.}
The normal equations of the inner problem give $\sum_{i\in\Msk} q_i = 0$ and
$\sum_{i\in\Msk} q_i \hat f_i = 0$, so a least-squares fit of $q$ on $\mathrm{span}\{1,\hat f\}$
alone returns zero. This is why \eqref{eq:trend} starts from $\hat f^2$ and $\log\hat f$; adding
$\hat f$ to them does change the joint fit, and App.~\ref{sec:appx-srp} measures that variant.

\FloatBarrier
\section{Method Details}
\label{sec:appx-method}

\subsection{Semiparametric residual propagation}

\paragraph{The linear system.} Problem \eqref{eq:harm} is assembled on the full-resolution grid over the four-neighbor graph, with
$\Aff$ the matrix of affinities \eqref{eq:rank}, $\Deg$ its row sums and $\Lap = \Deg - \Aff$
(we avoid $L$, $D$ and $W$, which denote the loss, the adapter dimension and a frozen weight
matrix). The measured pixels are eliminated rather than penalized, so the solved system is
$\Lap_{UU}\,\delta_U = -\,(\Lap\,\res)|_U$ with $\delta_\Msk = \res$ substituted in. By
Prop.~\ref{prop:harm} it is symmetric positive definite exactly when every pixel of $U$ has a
positive-weight path to $\Msk$. The affinity \eqref{eq:rank} assigns weight zero only to the globally
sharpest edge of each of its two factors, so at most two of the roughly $2N$ edges are cut;
two cut edges can disconnect a four-neighbor grid only by isolating a corner pixel incident to
both, a coincidence of probability zero for continuous gradients. We solve the system with
conjugate gradients preconditioned by the diagonal of $\Lap_{UU}$.

\paragraph{The cascadic hierarchy.}
Coarse levels are built by max-pooling the constraint mask---a coarse cell is measured if any
of its fine cells is, and takes the mean of those measurements---and by area-averaging the
edge weights. Every level runs the same fixed number of conjugate-gradient iterations on its
own grid and is prolonged bilinearly to the next, so coarse levels supply a starting point and
never an operator the fine level has to trust. The deployed budget is $4$ levels of
$120$ iterations.

\paragraph{The parametric term.} The fit \eqref{eq:trend} is a $3\times3$ normal-equation solve in double precision, fitted on the
measured pixels before the field is solved and added back afterward. It is evaluated at $\hat f$
clamped to the range $\hat f$ takes on $\Msk$, so the $\hat f^2$ and $\log\hat f$ terms never
extrapolate past depths the fit has seen.

\paragraph{Graph capture.}
Because the number of iterations at every level is fixed by the resolution alone, the whole
solve is a static computation graph: we capture it once per input shape as a CUDA graph and
replay it, which removes the per-iteration kernel-launch overhead that otherwise dominates its
cost (App.~\ref{sec:appx-results}). On every sample of the calibration set, we verify that the captured and uncaptured solutions agree bitwise.

\subsection{Spectral warm start}

Alg.~\ref{alg:sws} summarizes \sws{}. The gradient $G_l$ of every adapted layer comes from one
backward pass at the frozen weights, and the amplitude from forward passes along the ray, all
charged once per image; the deployed configuration uses $4$ forward passes in total
(Tab.~\ref{tab:amplitude}).

\begin{algorithm}[!htbp]
\caption{Spectral warm start (\sws{}) for one image.}
\label{alg:sws}
\begin{algorithmic}[1]
\Require frozen model with adapted layers $\{W_l\}$, measurements $y$ on $\Msk$, rank $r$, number
  of refinements $n$ (deployed: $n = 2$)
\State $G_l \gets \nabla_{W_l}\loss$ for every adapted layer \Comment{one backward pass}
\State $V_l \gets -\Fnorm{W_l}\,\proj_r(G_l)/\Fnorm{\proj_r(G_l)}$ \Comment{direction of \eqref{eq:sws}}
\State $u \gets \hat f|_\Msk$ at $\tau = 0$; \ $v \gets$ finite-difference derivative of $\hat f|_\Msk$
  along $\{V_l\}$ \Comment{one probe}
\State $\tau \gets \tau^\star(u, v, y)$ by \eqref{eq:tau}, or $0$ if its denominator is zero
\For{$k = 1, \dots, n$}
  \State evaluate $\hat f|_\Msk$ at $\tau$; \ $(u, v) \gets$ value at $\tau$ and secant slope
    through the last two evaluated points
  \State $\tau \gets \tau + \tau^\star(u, v, y)$ \Comment{same model, re-solved where it lands}
\EndFor
\State factor $\tau V_l$ into $sAB$ from its truncated singular vectors; free both factors
\State \Return the initialized adapter; run the unchanged $T$-step adaptation, then \srp{}
\end{algorithmic}
\end{algorithm}

\paragraph{One guard on \eqref{eq:tau}.}
The denominator of \eqref{eq:tau} vanishes exactly when the first-order model has no maximizer
along the ray (Prop.~\ref{prop:tau}). The implementation then takes $\tau = 0$, the frozen
initialization, rather than the unbounded step the tangent would license, so a degenerate ray
costs the run its warm start and nothing else. It is the only branch in \sws{} and not a tuned
threshold: the test is exact equality to zero in single precision.

\FloatBarrier
\section{Experimental Setup}
\label{sec:appx-setup}

\paragraph{Datasets and sparse inputs.}
Tab.~\ref{tab:sparsity} reports the measured set $\Msk$ per benchmark. The sparse inputs follow
the protocol of \citet{seo2026efficient} and differ in kind rather than only in density: NYUv2
and iBims-1 are uniform samples of a dense sensor, KITTI and DDAD are raw lidar returns
projected into the image, and VOID's points are whatever its visual-inertial front end tracks.
This is why VOID alone spans more than an order of magnitude in $m$ within one benchmark, and
why its measured set concentrates on corners and textured structures (App.~\ref{sec:appx-limits}).
Across all 6{,}479 images $m$ runs from $82$ to $18{,}966$, a factor of $231$.

\begin{table}[!htbp]
\centering
\caption{\textbf{The measured set $\Msk$ per benchmark.} Over all 6{,}479 evaluated images,
$m$ is the number of measured pixels and the share is $m$ over the image's pixel count.}
\label{tab:sparsity}
\footnotesize
\fittable{\begin{tabular}{lccccc}
\toprule
Benchmark & Images & \multicolumn{2}{c}{measured pixels $m$} & \multicolumn{2}{c}{share of the image} \\
\cmidrule(lr){3-4}\cmidrule(lr){5-6}
 & & median & range & median & range \\
\midrule
iBims-1 & 95 & 974 & 961--993 & 0.32\% & 0.32--0.33\% \\
VOID & 795 & 702 & 82--1,373 & 0.23\% & 0.03--0.46\% \\
NYUv2 & 649 & 475 & 455--488 & 0.72\% & 0.69--0.74\% \\
KITTI & 995 & 16,675 & 12,030--18,966 & 3.96\% & 2.85--4.50\% \\
DDAD & 3{,}945 & 4,782 & 1,989--5,788 & 0.47\% & 0.20--0.57\% \\
\bottomrule
\end{tabular}
}
\end{table}

\paragraph{Splits and the calibration set.}
The 6{,}479 evaluated images are 95 iBims-1, 795 VOID, 649 NYUv2, 995
KITTI and 3{,}945 DDAD, every image of every split. An earlier version of
the amplitude rule, reported as an ablation in Tab.~\ref{tab:amplitude}, placed its grid on a
calibration set drawn from the same five sources; several diagnostics also use it. It was drawn
to exclude the evaluation keys of a smaller evaluation set, so once every image of every split
is evaluated the two overlap, and we remove the 25 calibration images from the evaluated
set before any number is computed (6{,}479 of 6{,}504). The final configuration has no
per-benchmark or per-image constant, and nothing in it is calibrated at test time. Its discrete
design choices were compared on development subsets drawn from the evaluation splits.
Tab.~\ref{tab:decisions} lists each one, how it was decided and how it fares on held-out images;
App.~\ref{sec:appx-heldout} repeats Tab.~\ref{tab:main} on the 4{,}907 images outside every
development subset and repeats each comparison with one benchmark held out.

\begin{table}[!htbp]
\centering
\caption{\textbf{Every design decision, how it was made and how it holds out.} Development
subsets are evenly spaced draws from the evaluation splits; the held-out checks are in
App.~\ref{sec:appx-heldout}. ``Within $0.6$'' means that no alternative, re-selected with any one
benchmark held out, is better by more than $0.6$ points of relative MAE.}
\label{tab:decisions}
\footnotesize
\begin{tabular}{@{}>{\raggedright\arraybackslash}p{0.26\linewidth}>{\raggedright\arraybackslash}p{0.24\linewidth}>{\raggedright\arraybackslash}p{0.19\linewidth}>{\raggedright\arraybackslash}p{0.19\linewidth}@{}}
\toprule
Decision & How it was decided & Development subset & Held-out check \\
\midrule
Measurements as constraints, \eqref{eq:harm} & derived from (C1); ablated & 895 images (Tab.~\ref{tab:solve}) & unchanged \\
Rank affinity, \eqref{eq:rank} & derived from (C1); ablated & 895 images (Tab.~\ref{tab:solve}) & unchanged \\
Per-layer scale $\Fnorm{W_l}$, \eqref{eq:sws} & derived (Sec.~\ref{sec:sws}); ablated & 247 images (Tab.~\ref{tab:sws-ablation}) & unchanged \\
Parametric term, \eqref{eq:trend} & compared & 895 images (Tab.~\ref{tab:solve}) & within $0.6$ \\
Basis $[1, \hat f^2, \log \hat f]$ & order argument; size compared & 495 images (Tab.~\ref{tab:basis}) & within $0.6$ \\
Amplitude rule, $2$ refinements & compared & 495 images (Tab.~\ref{tab:amplitude}) & within $0.6$ \\
$10$-step operating point & smallest budget at the baseline in every cell & 895 images (Tab.~\ref{tab:budget}) & unchanged; holds on 4{,}907 held-out images \\
Cascadic budget $4\times120$ & cost choice; the converged solve is $1.1\%$ better & full splits, post hoc (Tab.~\ref{tab:solver}) & --- \\
Rank $8$, AdamW, lr $10^{-2}$, $40$ steps & \lora{}'s released configuration & --- & --- \\
\bottomrule
\end{tabular}
\end{table}

\paragraph{Metrics and statistics.}
MAE (mean absolute error) and RMSE (root-mean-square error) are $\frac{1}{|\mathcal{G}|}\sum_{i \in \mathcal{G}} |\hat d_i - y^{\mathrm{gt}}_i|$ and
$\bigl(\frac{1}{|\mathcal{G}|}\sum_{i \in \mathcal{G}} (\hat d_i - y^{\mathrm{gt}}_i)^2\bigr)^{1/2}$
in meters, with $\hat d$ the dense output and $\mathcal{G}$ the pixels with valid ground truth;
AbsRel (absolute relative error) is the mean of $|\hat d_i - y^{\mathrm{gt}}_i|/y^{\mathrm{gt}}_i$, and $\delta_1$ the fraction of
pixels with $\max(\hat d_i/y^{\mathrm{gt}}_i,\, y^{\mathrm{gt}}_i/\hat d_i) < 1.25$. Every benchmark mean
weights its images equally. Relative figures are signed so that negative is an improvement;
aggregates are unweighted means of per-benchmark relatives unless stated as pooled per image,
in which case confidence intervals and $p$-values come from a paired bootstrap over images.

\paragraph{Baselines.}
The foundation-model rows of Tab.~\ref{tab:main} run each backbone's released model on the same
inputs, raw and after a least-squares scale-and-shift fit to the measurements.
\testprompt{}~\citep{jeong2025testprompt} is re-implemented and run under our protocol.
\lora{}~\citep{seo2026efficient} uses its released configuration (Tab.~\ref{tab:hparams}) and is
averaged over five seeds. Marigold-DC~\citep{viola2025marigolddc} runs with its official
code and settings---50 denoising steps, one sample, bf16, and the processing resolution of the
authors' benchmark scripts (native for iBims-1, VOID and KITTI; 768 pixels for NYUv2 and
DDAD)---on the same inputs and masks, and is scored on its own output without re-alignment.
Its 10-sample ensemble and four further seeds run on a subset fixed before any result was seen:
all 95 iBims-1 images and 40 per other benchmark by a fixed seed (App.~\ref{sec:appx-more}).
\citet{lee2025zeroshotdc} is not run.

\begin{table}[!htbp]
\centering
\caption{\textbf{Configuration.} Everything not listed is inherited unchanged from the released
\lora{} configuration~\citep{seo2026efficient}; neither component adds a constant in data units.}
\label{tab:hparams}
\footnotesize
\begin{tabular}{@{}ll>{\raggedright\arraybackslash}p{0.64\linewidth}@{}}
\toprule
Component & Setting & Value \\
\midrule
Backbone & frozen model & UniDepthV2 with a ViT-L encoder~\citep{piccinelli2025unidepthv2} \\
 & second backbone & MoGe-2~\citep{wang2025moge2}, in Tab.~\ref{tab:backbone} \\
Adapter & placement, rank & decoder linear layers (MoGe-2: neck and heads), $r = 8$ \\
 & parameters & $D = 438{,}784$ (UniDepthV2) \\
Adaptation & optimizer & AdamW~\citep{loshchilov2019adamw}, learning rate $10^{-2}$ \\
 & weight decay & as released \\
 & steps & $10$ or $40$ (\lora{}: $40$) \\
 & seeds & five for zero-initialized runs; \sws{} is deterministic \\
\sws{} & direction & rank-$r$ truncated SVD of $G_l$, scaled by $\Fnorm{W_l}$ \eqref{eq:sws} \\
 & amplitude & \eqref{eq:tau}, $2$ secant refinements; $1$ backward and $4$ forward passes \\
\srp{} & graph, affinity & four-neighborhood at full resolution; ranks \eqref{eq:rank}, no bandwidth \\
 & parametric fit & $[1, \hat f^2, \log\hat f]$ \eqref{eq:trend}, double precision, $\hat f$ clamped to its range on $\Msk$ \\
 & solver & Jacobi-preconditioned CG, cascadic, $4\times 120$ iterations; CUDA graph \\
Hardware & timing & one NVIDIA RTX 4090 \\
\bottomrule
\end{tabular}
\end{table}

\paragraph{Timing protocol.}
Every latency is measured by us on one NVIDIA RTX 4090, each configuration in its own process:
10 warm-up images, then 30 timed images, reported as the median, with regions bracketed by
\texttt{cuda.synchronize}. The nominal resolutions VGA, SD, HD and FHD are cropped to a multiple
of the backbone's patch size, giving $476\times630$, $476\times714$, $714\times1274$ and $1078\times1918$.
Marigold-DC, one sample of 50 denoising steps in bf16, is timed at its default
processing resolution, 768 pixels on the long side, which is also its authors' setting for NYUv2;
every input is resized to it, so its latency barely depends on the input's size. At the input's
own resolution it is faster at VGA and SD, slower at HD, and does not fit in 24~GB at FHD
(Tab.~\ref{tab:cost}).

\paragraph{Determinism.}
The forward pass is bitwise reproducible, and backward reductions are made deterministic, which
the seed-variance measurements of Tab.~\ref{tab:seed} require: without it, repeated runs with
the same seed would differ at the level being measured.

\FloatBarrier
\section{Additional Results}
\label{sec:appx-results}

\paragraph{All four metrics.}
Tab.~\ref{tab:full} repeats Tab.~\ref{tab:main} with AbsRel and $\delta_1$. AbsRel tracks MAE
closely everywhere, and $\delta_1$ is near saturation on every benchmark except DDAD, where the
frozen backbone's scale error is large enough to push pixels past the threshold. The ordering of
the configurations is the same as in Tab.~\ref{tab:main}; in particular, the $10$-step
configuration matches or beats \lora{} on all four metrics on every benchmark.

\begin{table}[!htbp]
\centering
\caption{\textbf{All four metrics for the rows of Tab.~\ref{tab:main}.} Same runs and
6{,}479 images; the step count is in parentheses.}
\label{tab:full}
\footnotesize
\fittable{\begin{tabular}{llcccc}
\toprule
Benchmark & Method (steps) & MAE $\downarrow$ & RMSE $\downarrow$ & AbsRel $\downarrow$ & $\delta_1$ $\uparrow$ \\
\midrule
\multirow{5}{*}{iBims-1} & $+$ \lora{}~\citep{seo2026efficient} (40) & 0.041 & 0.146 & 0.011 & 0.994 \\
 & $+$ \textbf{\srp{}} \emph{(ours)} (0) & 0.040 & 0.162 & 0.011 & 0.992 \\
 & $+$ \textbf{\sws{} $+$ \srp{}} \emph{(ours)} (10) & 0.041 & 0.144 & 0.011 & 0.994 \\
 & $+$ \lora{} $+$ \srp{} \emph{(ours)} (40) & 0.037 & 0.146 & 0.010 & 0.994 \\
 & $+$ \sws{} $+$ \srp{} \emph{(ours)} (40) & 0.037 & 0.146 & 0.010 & 0.994 \\
\midrule
\multirow{5}{*}{VOID} & $+$ \lora{}~\citep{seo2026efficient} (40) & 0.136 & 0.505 & 0.026 & 0.982 \\
 & $+$ \textbf{\srp{}} \emph{(ours)} (0) & 0.152 & 0.539 & 0.030 & 0.975 \\
 & $+$ \textbf{\sws{} $+$ \srp{}} \emph{(ours)} (10) & 0.134 & 0.504 & 0.025 & 0.982 \\
 & $+$ \lora{} $+$ \srp{} \emph{(ours)} (40) & 0.132 & 0.503 & 0.023 & 0.982 \\
 & $+$ \sws{} $+$ \srp{} \emph{(ours)} (40) & 0.130 & 0.501 & 0.022 & 0.983 \\
\midrule
\multirow{5}{*}{NYUv2} & $+$ \lora{}~\citep{seo2026efficient} (40) & 0.054 & 0.125 & 0.019 & 0.992 \\
 & $+$ \textbf{\srp{}} \emph{(ours)} (0) & 0.055 & 0.142 & 0.020 & 0.989 \\
 & $+$ \textbf{\sws{} $+$ \srp{}} \emph{(ours)} (10) & 0.054 & 0.125 & 0.019 & 0.992 \\
 & $+$ \lora{} $+$ \srp{} \emph{(ours)} (40) & 0.050 & 0.123 & 0.017 & 0.992 \\
 & $+$ \sws{} $+$ \srp{} \emph{(ours)} (40) & 0.049 & 0.122 & 0.017 & 0.992 \\
\midrule
\multirow{5}{*}{KITTI} & $+$ \lora{}~\citep{seo2026efficient} (40) & 0.438 & 1.372 & 0.023 & 0.993 \\
 & $+$ \textbf{\srp{}} \emph{(ours)} (0) & 0.340 & 1.448 & 0.017 & 0.991 \\
 & $+$ \textbf{\sws{} $+$ \srp{}} \emph{(ours)} (10) & 0.321 & 1.301 & 0.016 & 0.993 \\
 & $+$ \lora{} $+$ \srp{} \emph{(ours)} (40) & 0.318 & 1.325 & 0.015 & 0.993 \\
 & $+$ \sws{} $+$ \srp{} \emph{(ours)} (40) & 0.310 & 1.329 & 0.015 & 0.994 \\
\midrule
\multirow{5}{*}{DDAD} & $+$ \lora{}~\citep{seo2026efficient} (40) & 2.002 & 5.937 & 0.061 & 0.962 \\
 & $+$ \textbf{\srp{}} \emph{(ours)} (0) & 1.480 & 6.325 & 0.041 & 0.968 \\
 & $+$ \textbf{\sws{} $+$ \srp{}} \emph{(ours)} (10) & 1.386 & 5.494 & 0.037 & 0.971 \\
 & $+$ \lora{} $+$ \srp{} \emph{(ours)} (40) & 1.399 & 5.613 & 0.037 & 0.972 \\
 & $+$ \sws{} $+$ \srp{} \emph{(ours)} (40) & 1.341 & 5.695 & 0.036 & 0.974 \\
\bottomrule
\end{tabular}
}
\end{table}

\paragraph{Paired statistics for the $10$-step configuration.}
Sec.~\ref{sec:main} gives paired statistics for the equal-budget row; the configuration we lead
with receives the same treatment here. Averaging per-benchmark relatives, \sws{} at $10$ steps
read out through \srp{} changes MAE by $-15.1\%$ over the core and
$-12.3\%$ over all five; pooled per image it gives $-25.8\%$
($[-26.2,\,-25.5]$, $p\,<10^{-4}$), improving $94.2\%$ of the core's images. The pooled
figure is larger for an arithmetic reason: DDAD is $69\%$ of the core pool and has the
largest gain. Weighting the per-benchmark relatives by image count already gives
$-26.2\%$ against the pooled $-25.8\%$, and $-26.7\%$ against
$-26.5\%$ for the equal-budget row, so the imbalance accounts for essentially the whole
difference. We therefore quote the unweighted per-benchmark figure as the headline throughout.

\paragraph{Win rates per metric.}
For \srp{} added to the $40$-step run, MAE improves on $100.0\%$ of images over the
core and $99.9\%$ over all five, and RMSE on $96.4\%$ and
$95.6\%$. The two smallest benchmark-level gains are VOID RMSE
($0.33\%$) and iBims-1 RMSE ($0.47\%$). At $10$ steps the rates are
$90.6\%$ on MAE and $80.7\%$ on RMSE over all five. RMSE is lower
than MAE in both cases because a squared metric is priced by the pixels the field reaches
least well, and on a minority of images that tail moves the wrong way even where the benchmark
mean improves; 288 of the 6{,}479 images end with a higher RMSE at the
full budget, most often on VOID (11\% of its images).

\paragraph{Density and remaining error.}
One might suspect that the gain is purely a density effect. Over the core, the Spearman
correlation between a sample's relative improvement and its measurement density is $+0.34$
($p\,<10^{-4}$): since improvements are negative, the gain shrinks as measurements get denser.
The correlation with the baseline's error on that sample is stronger, at $-0.50$: the gain is
largest where the most error remains, whatever the density. This is the pattern
Sec.~\ref{sec:manifold} predicts, since the choice among minimizers is present on every sample,
not only on sparse ones.

\paragraph{The design choices hold out.}
\label{sec:appx-heldout}
Tab.~\ref{tab:decisions} lists every design decision and the development subset on which it was
compared. Tab.~\ref{tab:heldout} repeats Tab.~\ref{tab:main} on the 4{,}907 images that lie outside
every development subset (430 VOID, 310 NYUv2, 620 KITTI and 3{,}547 DDAD; each of the 95 iBims-1
images enters at least one subset). Every conclusion of Tab.~\ref{tab:main} holds there: the
$10$-step configuration matches or beats the $40$-step baseline in all eight cells, both
equal-budget rows improve every cell, and the per-benchmark changes in MAE agree with the full
splits to within two points ($-1.3\%$, $-1.0\%$, $-27.3\%$ and $-30.7\%$ for the $10$-step
configuration, against $-1.1\%$, $-1.4\%$, $-26.8\%$ and $-30.7\%$).

Tab.~\ref{tab:lobo} repeats each comparison with one benchmark held out, by the criterion of its
own table. The constraint, the rank affinity, the warm-start variant and the $10$-step budget are
re-selected unchanged in every case, and every other choice stays within $0.6$ points of its best
alternative; at $40$ steps all of them are re-selected unchanged.
Swapping each choice for its neighbour, on the images it was compared on, moves the mean MAE of
the configurations that use \srp{} by at most $0.6\%$: zero, one or three refinements by $+0.33\%$,
$+0.37\%$ and $+0.22\%$ at $40$ steps ($+0.41\%$, $+0.09\%$ and $+0.16\%$ at $10$), the best greedy grid by $+0.57\%$ and
$+0.33\%$, and a four-term basis by $+0.52\%$ without adaptation. Without the read-out the
refinements matter more, since no refinement costs $4.0\%$ at $10$ steps, which is why the
deployed rule keeps two.

\begin{table}[!htbp]
\centering
\caption{\textbf{Tab.~\ref{tab:main} on the 4{,}907 held-out images.} The same runs as
Tab.~\ref{tab:main}, restricted to the images outside every development subset of
Tab.~\ref{tab:decisions}; iBims-1 has none. MAE and RMSE in meters.}
\label{tab:heldout}
\footnotesize
\fittable{\begin{tabular}{lccccccccc}
\toprule
 & & \multicolumn{2}{c}{VOID (430)} & \multicolumn{2}{c}{NYUv2 (310)} & \multicolumn{2}{c}{KITTI (620)} & \multicolumn{2}{c}{DDAD (3{,}547)} \\
\cmidrule(lr){3-4}\cmidrule(lr){5-6}\cmidrule(lr){7-8}\cmidrule(lr){9-10}
Method & Steps & MAE & RMSE & MAE & RMSE & MAE & RMSE & MAE & RMSE \\
\midrule
Frozen decoder $+$ scale\,\&\,shift & 0 & 0.161 & 0.476 & 0.161 & 0.263 & 1.428 & 3.030 & 5.905 & 14.128 \\
$+$ \lora{} & 40 & 0.106 & 0.411 & 0.053 & 0.122 & 0.441 & 1.376 & 2.003 & 5.956 \\
$+$ \sws{} & 40 & 0.104 & 0.408 & 0.051 & 0.121 & 0.399 & 1.360 & 1.754 & 5.866 \\
\midrule
$+$ \textbf{\srp{}} & \textbf{0} & 0.121 & 0.444 & 0.053 & 0.139 & 0.340 & 1.454 & 1.483 & 6.344 \\
$+$ \textbf{\sws{} $+$ \srp{}} & \textbf{10} & 0.105 & 0.410 & 0.052 & 0.122 & 0.321 & 1.303 & 1.389 & 5.511 \\
$+$ \lora{} $+$ \srp{} & 40 & 0.103 & 0.409 & 0.049 & 0.120 & 0.318 & 1.328 & 1.402 & 5.632 \\
$+$ \sws{} $+$ \srp{} & 40 & 0.101 & 0.407 & 0.048 & 0.120 & 0.310 & 1.333 & 1.343 & 5.716 \\
\bottomrule
\end{tabular}
}
\end{table}

\begin{table}[!htbp]
\centering
\caption{\textbf{Each comparison repeated with one benchmark held out.} The criterion is that of
the table the decision comes from, the mean relative MAE over the remaining benchmarks. ``$=$'':
the deployed choice is re-selected; otherwise the winner and its margin in points.}
\label{tab:lobo}
\footnotesize
\fittable{\begin{tabular}{llccccc}
\toprule
 & & \multicolumn{5}{c}{benchmark held out} \\
\cmidrule(lr){3-7}
Decision & Compared at & iBims-1 & VOID & NYUv2 & KITTI & DDAD \\
\midrule
Constraint, rank affinity & $0$ and $40$ steps & $=$ & $=$ & $=$ & $=$ & $=$ \\
Parametric term & $0$ steps & $=$ & $=$ & $=$ & $=$ & dropped ($0.24$) \\
 & $40$ steps & $=$ & $=$ & $=$ & $=$ & $=$ \\
Basis $[1, \hat f^2, \log \hat f]$ & $0$ steps & $=$ & $=$ & $=$ & $=$ & position ($0.57$) \\
Amplitude rule, $2$ refinements & $10$ steps & greedy grid ($0.11$) & $=$ & $=$ & $=$ & $=$ \\
 & $40$ steps & $=$ & $=$ & $=$ & $=$ & $=$ \\
Warm-start variant & $10$ and $40$ steps & $=$ & $=$ & $=$ & $=$ & $=$ \\
$10$-step operating point & budget sweep & $=$ & $=$ & $=$ & $=$ & $=$ \\
\bottomrule
\end{tabular}
}
\end{table}

\paragraph{A second backbone.}
Tab.~\ref{tab:backbone} repeats the leading rows on UniDepthV2 and MoGe-2, which differ both in
head (a pixel decoder against a point-map head) and in how they obtain metric scale. Nothing is
transferred and nothing is retuned: \eqref{eq:sws} reads each model's own gradient at its own
frozen weights, and \eqref{eq:tau} is re-solved on its own ray. UniDepthV2 is re-run on the same
250 images as MoGe-2, so the two blocks form one comparison. At equal budget,
\sws{}~$+$~\srp{} changes MAE against each backbone's own $40$-step \lora{} by
$-20.4\%$ and $-15.6\%$ over the core ($-16.7\%$ and
$-12.8\%$ over all five), wins all $4$ driving cells on both,
and is behind on $2$ and $0$ of the $6$ indoor
cells. With no backward pass, \srp{} changes MAE against a full $40$-step adaptation of the same model by $-21.3\%$ and $-24.0\%$ on KITTI and by $-27.3\%$ and
$-25.4\%$ on DDAD. At $10$ steps the driving gains remain ($-25.4\%$
and $-31.4\%$ on UniDepthV2, $-22.7\%$ and $-23.0\%$ on
MoGe-2), while $3$ and $2$ of the $6$ indoor cells
move the wrong way, by at most $+1.1\%$ (VOID RMSE) and
$+0.8\%$ (VOID RMSE). On these subsets, the $10$-step point
trades a small indoor cost for large driving gains, and the equal-budget row removes the
indoor cost.

\begin{table}[!htbp]
\centering
\caption{\textbf{The leading rows of Tab.~\ref{tab:main} on two frozen backbones.} The adapter
sits in whatever follows each frozen encoder, and nothing is retuned. \best{Bold and underlined}
marks the best value in each column of each block. Absolute errors are comparable within a block
but not across blocks, because each model predicts on its own grid. Both blocks use the same
250 images (50 iBims-1, 50 VOID, 50 NYUv2, 50 KITTI and 50 DDAD).}
\label{tab:backbone}
\footnotesize
\fittable{\begin{tabular}{lccccccccccc}
\toprule
 &  & \multicolumn{2}{c}{iBims-1} & \multicolumn{2}{c}{VOID} & \multicolumn{2}{c}{NYUv2} & \multicolumn{2}{c}{KITTI} & \multicolumn{2}{c}{DDAD} \\
\cmidrule(lr){3-4}\cmidrule(lr){5-6}\cmidrule(lr){7-8}\cmidrule(lr){9-10}\cmidrule(lr){11-12}
Method & Steps & MAE & RMSE & MAE & RMSE & MAE & RMSE & MAE & RMSE & MAE & RMSE \\
\midrule
\multicolumn{12}{l}{\emph{UniDepthV2~\citep{piccinelli2025unidepthv2}}} \\
\quad frozen decoder $+$ scale\,\&\,shift & 0 & 0.099 & 0.226 & 0.262 & 0.739 & 0.170 & 0.282 & 1.518 & 3.142 & 6.230 & 14.521 \\
\quad $+$ \lora{} & 40 & 0.043 & 0.154 & 0.208 & \best{0.675} & 0.054 & 0.128 & 0.453 & 1.406 & 2.057 & 6.140 \\
\quad $+$ \textbf{\srp{}} \emph{(ours)} & \textbf{0} & 0.042 & 0.170 & 0.216 & 0.696 & 0.055 & 0.147 & 0.356 & 1.504 & 1.496 & 6.486 \\
\quad $+$ \textbf{\sws{} $+$ \srp{}} \emph{(ours)} & \textbf{10} & 0.042 & \best{0.150} & 0.209 & 0.682 & 0.053 & 0.129 & 0.338 & \best{1.353} & 1.411 & \best{5.650} \\
\quad $+$ \sws{} $+$ \srp{} \emph{(ours)} & 40 & \best{0.039} & 0.156 & \best{0.204} & 0.676 & \best{0.049} & \best{0.127} & \best{0.325} & 1.377 & \best{1.358} & 5.894 \\
\midrule
\multicolumn{12}{l}{\emph{MoGe-2~\citep{wang2025moge2}}} \\
\quad frozen decoder $+$ scale\,\&\,shift & 0 & 0.093 & 0.204 & 0.262 & 0.734 & 0.138 & 0.245 & 1.243 & 2.720 & 3.717 & 9.064 \\
\quad $+$ \lora{} & 40 & 0.042 & 0.148 & 0.206 & 0.670 & 0.055 & 0.122 & 0.439 & 1.421 & 1.812 & 6.109 \\
\quad $+$ \textbf{\srp{}} \emph{(ours)} & \textbf{0} & \best{0.037} & 0.155 & 0.212 & 0.688 & \best{0.050} & 0.132 & 0.333 & 1.395 & \best{1.351} & 5.869 \\
\quad $+$ \textbf{\sws{} $+$ \srp{}} \emph{(ours)} & \textbf{10} & 0.040 & \best{0.142} & 0.206 & 0.676 & 0.052 & 0.120 & 0.339 & \best{1.357} & 1.395 & \best{5.762} \\
\quad $+$ \sws{} $+$ \srp{} \emph{(ours)} & 40 & \best{0.037} & \best{0.142} & \best{0.202} & \best{0.668} & 0.051 & \best{0.119} & \best{0.332} & 1.377 & 1.448 & 6.081 \\
\bottomrule
\end{tabular}
}
\end{table}

\paragraph{The budget axis.}
Tab.~\ref{tab:budget} sweeps the adaptation budget for both initializations, with and without
the read-out. \textbf{The read-out substitutes for most of the adaptation}: $2$ adaptation steps
read out through \srp{} reach 92\% of the $40$-step baseline's MAE over the core,
against 227\% without it. The gain is concentrated on the driving benchmarks, and $2$
steps leave the RMSE tail behind almost everywhere; closing it is what further steps buy.
\textbf{The warm start makes the short budget usable}: at $10$ steps with the read-out, the
zero initialization beats the baseline in $5$ of the $10$ benchmark--metric cells,
while \sws{} matches or beats it in all $10$; without the read-out, \sws{} reaches the
baseline's full-budget accuracy in $30$ steps. The steps a zero-initialized run spends
leaving the degenerate point of (C3) are a fixed cost, so removing them matters most when the
budget is tightest.

\begin{table}[!htbp]
\centering
\caption{\textbf{The budget axis.} Both initializations at eight adaptation budgets, with and without
the read-out; rows differ only in the initialization, the number of steps and the read-out.
Bold rows mark the baseline's shipped budget and the configuration we lead with. 895
images, up to 200 per benchmark.}
\label{tab:budget}
\footnotesize
\fittable{\begin{tabular}{lcccccccccccc}
\toprule
 & \multicolumn{2}{c}{iBims-1} & \multicolumn{2}{c}{VOID} & \multicolumn{2}{c}{NYUv2} & \multicolumn{2}{c}{KITTI} & \multicolumn{2}{c}{DDAD} \\
\cmidrule(lr){2-3}\cmidrule(lr){4-5}\cmidrule(lr){6-7}\cmidrule(lr){8-9}\cmidrule(lr){10-11}
Init.\ / steps & MAE & RMSE & MAE & RMSE & MAE & RMSE & MAE & RMSE & MAE & RMSE \\
\midrule
\multicolumn{11}{l}{\emph{the adaptation alone}} \\
\lora{}, 2 & 0.092 & 0.192 & 0.220 & 0.720 & 0.155 & 0.250 & 0.961 & 2.217 & 3.591 & 9.926 \\
\lora{}, 5 & 0.076 & 0.170 & 0.207 & 0.694 & 0.117 & 0.192 & 0.890 & 1.951 & 3.518 & 7.955 \\
\lora{}, 10 & 0.061 & 0.155 & 0.194 & 0.675 & 0.089 & 0.157 & 0.692 & 1.661 & 2.815 & 6.986 \\
\lora{}, 15 & 0.055 & 0.152 & 0.186 & 0.671 & 0.075 & 0.140 & 0.588 & 1.506 & 2.554 & 6.501 \\
\lora{}, 20 & 0.049 & 0.149 & 0.182 & 0.668 & 0.067 & 0.133 & 0.534 & 1.445 & 2.366 & 6.250 \\
\lora{}, 25 & 0.046 & 0.147 & 0.178 & 0.663 & 0.062 & 0.130 & 0.497 & 1.411 & 2.276 & 6.167 \\
\lora{}, 30 & 0.044 & 0.147 & 0.177 & 0.662 & 0.059 & 0.127 & 0.466 & 1.373 & 2.148 & 6.043 \\
\lora{}, 40 & \textbf{0.041} & \textbf{0.147} & \textbf{0.174} & \textbf{0.660} & \textbf{0.055} & \textbf{0.125} & \textbf{0.429} & \textbf{1.353} & \textbf{1.978} & \textbf{5.909} \\
\sws{}, 2 & 0.094 & 0.183 & 0.205 & 0.693 & 0.124 & 0.201 & 0.948 & 2.069 & 3.248 & 7.649 \\
\sws{}, 5 & 0.084 & 0.186 & 0.194 & 0.678 & 0.099 & 0.169 & 0.716 & 1.688 & 2.843 & 6.987 \\
\sws{}, 10 & 0.058 & 0.154 & 0.185 & 0.666 & 0.078 & 0.144 & 0.583 & 1.496 & 2.403 & 6.367 \\
\sws{}, 15 & 0.053 & 0.151 & 0.181 & 0.663 & 0.068 & 0.135 & 0.519 & 1.428 & 2.214 & 6.092 \\
\sws{}, 20 & 0.048 & 0.149 & 0.178 & 0.662 & 0.062 & 0.129 & 0.477 & 1.382 & 2.062 & 5.947 \\
\sws{}, 25 & 0.045 & 0.149 & 0.176 & 0.661 & 0.058 & 0.127 & 0.446 & 1.354 & 1.976 & 5.917 \\
\sws{}, 30 & 0.043 & 0.148 & 0.174 & 0.659 & 0.056 & 0.125 & 0.423 & 1.346 & 1.879 & 5.865 \\
\sws{}, 40 & 0.040 & 0.147 & 0.172 & 0.658 & 0.053 & 0.124 & 0.395 & 1.339 & 1.765 & 5.850 \\
\midrule
\multicolumn{11}{l}{\emph{$+$ \srp{}}} \\
no adaptation & 0.040 & 0.162 & 0.193 & 0.693 & 0.056 & 0.143 & 0.339 & 1.435 & 1.484 & 6.404 \\
\lora{}, 2 & 0.046 & 0.151 & 0.187 & 0.682 & 0.058 & 0.131 & 0.326 & 1.315 & 1.454 & 6.190 \\
\lora{}, 5 & 0.045 & 0.149 & 0.178 & 0.668 & 0.061 & 0.131 & 0.333 & 1.313 & 1.471 & 5.694 \\
\lora{}, 10 & 0.042 & 0.146 & 0.175 & 0.663 & 0.057 & 0.127 & 0.329 & 1.306 & 1.451 & 5.581 \\
\lora{}, 15 & 0.041 & 0.147 & 0.173 & 0.663 & 0.055 & 0.124 & 0.325 & 1.299 & 1.436 & 5.528 \\
\lora{}, 20 & 0.040 & 0.146 & 0.173 & 0.662 & 0.054 & 0.124 & 0.323 & 1.302 & 1.427 & 5.536 \\
\lora{}, 25 & 0.039 & 0.145 & 0.171 & 0.660 & 0.053 & 0.124 & 0.321 & 1.308 & 1.416 & 5.538 \\
\lora{}, 30 & 0.038 & 0.145 & 0.171 & 0.659 & 0.052 & 0.123 & 0.319 & 1.307 & 1.410 & 5.566 \\
\lora{}, 40 & 0.037 & 0.146 & 0.170 & 0.658 & 0.051 & 0.123 & 0.316 & 1.312 & 1.396 & 5.608 \\
\sws{}, 2 & 0.046 & 0.149 & 0.176 & 0.665 & 0.060 & 0.129 & 0.329 & 1.311 & 1.420 & 5.675 \\
\sws{}, 5 & 0.041 & 0.151 & 0.174 & 0.662 & 0.057 & 0.127 & 0.324 & 1.294 & 1.408 & 5.589 \\
\sws{}, 10 & \textbf{0.041} & \textbf{0.144} & \textbf{0.171} & \textbf{0.657} & \textbf{0.055} & \textbf{0.125} & \textbf{0.320} & \textbf{1.290} & \textbf{1.387} & \textbf{5.530} \\
\sws{}, 15 & 0.040 & 0.147 & 0.171 & 0.658 & 0.053 & 0.124 & 0.318 & 1.299 & 1.376 & 5.526 \\
\sws{}, 20 & 0.039 & 0.146 & 0.170 & 0.658 & 0.052 & 0.123 & 0.314 & 1.294 & 1.366 & 5.553 \\
\sws{}, 25 & 0.038 & 0.147 & 0.171 & 0.658 & 0.051 & 0.123 & 0.312 & 1.297 & 1.356 & 5.593 \\
\sws{}, 30 & 0.038 & 0.147 & 0.170 & 0.657 & 0.050 & 0.123 & 0.310 & 1.304 & 1.347 & 5.621 \\
\sws{}, 40 & 0.037 & 0.146 & 0.169 & 0.657 & 0.050 & 0.122 & 0.308 & 1.310 & 1.338 & 5.696 \\
\bottomrule
\end{tabular}
}
\end{table}

\paragraph{Latency.}
Tab.~\ref{tab:cost} breaks the latency down by component and resolution. Solving
\eqref{eq:harm} by plain conjugate gradients to convergence costs 1{,}425~ms at VGA;
the cascadic cycle comes within $1.1\%$ of that field (Tab.~\ref{tab:solver}), and
capturing it saves a further factor of 5--10. The uncaptured solve costs
0.45~ms per iteration at every resolution we tried, from $0.3$ to $2$ megapixels; a cost
that does not grow with the problem is launch overhead, which is exactly what capture removes.
At the adaptation-free point the whole pipeline runs at 9.2 frames per second at VGA and
1.0 at FHD. Capture is paid once per input shape: the first call, three eager passes and the
capture, takes $804$--$933$~ms from VGA to FHD and $27$--$177$~MiB of extra memory, after which every
input of that shape replays in $18.3$--$40.3$~ms (Tab.~\ref{tab:cost}).
Without capture, the same solve takes $179$--$184$~ms and returns the identical field.

\begin{table}[!htbp]
\centering
\caption{\textbf{Latency by input resolution.} Milliseconds per image, median (protocol in
App.~\ref{sec:appx-setup}). ``Steps'' is the number of adaptation steps per image. Indented rows
decompose the row above; italic rows are variants of the deployed solve that we do not ship.
The warm start's cost, one gradient evaluation and $4$ forward passes, is charged once
per image. Marigold-DC processes every input at 768 pixels, its default, or at the
input's own resolution, which at FHD exceeds the 24~GB of one RTX 4090 (---); $^\dagger$denoising
steps.}
\label{tab:cost}
\footnotesize
\fittable{\begin{tabular}{llrrrr}
\toprule
 & & \multicolumn{4}{c}{latency (ms) by input resolution} \\
\cmidrule(lr){3-6}
Method & Steps & VGA & SD & HD & FHD \\
\midrule
Frozen forward pass & 0 & 88 & 101 & 303 & 976 \\
\midrule
\testprompt{}~\citep{jeong2025testprompt} & --- & 6{,}793 & 8{,}404 & 34{,}035 & 134{,}346 \\
Marigold-DC~\citep{viola2025marigolddc} & 50$^\dagger$ & 21{,}957 & 19{,}055 & 17{,}222 & 17{,}211 \\
Marigold-DC, native resolution & 50$^\dagger$ & 16{,}358 & 18{,}020 & 43{,}338 & --- \\
\lora{}~\citep{seo2026efficient} & 40 & 4{,}343 & 4{,}869 & 12{,}361 & 30{,}354 \\
\sws{} \emph{(ours)} & 40 & 4{,}497 & 5{,}062 & 13{,}197 & 32{,}632 \\
\midrule
\textbf{\srp{} only} \emph{(ours)} & \textbf{0} & 109 & 123 & 331 & 1{,}024 \\
\quad affinity \eqref{eq:rank} & & 1.4 & 1.4 & 1.9 & 4.4 \\
\quad parametric fit \eqref{eq:trend} & & 1.3 & 1.3 & 1.8 & 2.9 \\
\quad cascadic solve, captured & & 18.3 & 18.8 & 24.1 & 40.3 \\
\quad \emph{same, not captured} & & 179 & 180 & 184 & 182 \\
\quad \emph{same, plain CG to convergence} & & 1{,}425 & 1{,}431 & 1{,}429 & 1{,}431 \\
\midrule
\textbf{\sws{} $+$ \srp{}} \emph{(ours)} & \textbf{10} & 1{,}728 & 1{,}916 & 4{,}551 & 11{,}149 \\
\sws{} $+$ \srp{} \emph{(ours)} & 40 & 4{,}517 & 5{,}083 & 13{,}225 & 32{,}680 \\
\bottomrule
\end{tabular}
}
\end{table}

\FloatBarrier
\section{Empirical Verification of the Analysis}
\label{sec:appx-verify}

The statements of Sec.~\ref{sec:analysis} are proved in App.~\ref{sec:appx-proofs}. What remains
to check is that the objective the code minimizes is the one analyzed, and that the effects the
three consequences predict are large enough to matter. Tab.~\ref{tab:analysis} does the first
on every sample of the diagnostic subset; the figures below do the second.

\begin{table}[!htbp]
\centering
\caption{\textbf{The closed form and its consequences.} Measured on the 50-image diagnostic
subset (range over samples). First block: the implemented objective against \eqref{eq:closed},
and the flat manifolds of (C2), the factorization one over scalar rescalings and over the full
group $\mathrm{GL}(r)$. Second block: the degenerate initialization (C3), with $A$ and $B$ the
up- and down-projections. Third block: the data scale (C1), the under-determination of
Sec.~\ref{sec:manifold}, and the submersion check (App.~\ref{sec:appx-verify}).}
\label{tab:analysis}
\footnotesize
\setlength{\tabcolsep}{4pt}
\fittable{\begin{tabular}{llc}
\toprule
Property & Quantity & Measured over the sample set \\
\midrule
Closed form \eqref{eq:closed} & max.\ rel.\ deviation from the implemented loss & $8.1 \times 10^{-8}$ -- $1.7 \times 10^{-6}$ \\
Envelope identity & $\cos(\nabla\loss_{\mathrm{detached}}, \nabla\loss_{\mathrm{exact}})$ & 0.999984 -- 1.000000 \\
Output-affine invariance & max.\ rel.\ deviation of $\loss$ & $0$ -- $1.1 \times 10^{-6}$ \\
Factor rescale invariance & max.\ rel.\ deviation of $\loss$ & $1.4 \times 10^{-6}$ -- $8.0 \times 10^{-4}$ \\
\quad the full gauge group & the same, under $(A,B)\mapsto(AM,M^{-1}B)$ & $1.7 \times 10^{-5}$ -- $5.0 \times 10^{-3}$ \\
\midrule
Degenerate init.\ (C3) & fraction of samples with $\nabla_{\!B}\loss \equiv 0$ & 50/50 \\
 & fraction of negative Ritz values & 0.47 -- 0.50 \\
 & $\norm{\nabla_{\!A}\loss}/\norm{\nabla_{\!B}\loss}$ near init.\ & $3.2 \times 10^{4}$ -- $2.4 \times 10^{5}$ \\
\midrule
Data scale (C1) & $\loss$ at initialization & $1.0 \times 10^{-3}$ -- $6.4 \times 10^{2}$ \\
 & $1-\rho_\Msk^2$ at initialization & $3.6 \times 10^{-3}$ -- $4.9 \times 10^{-1}$ \\
Under-determination & measured pixels $m$ & 419 -- 17{,}923 \\
 & $1-\rho_\Msk^2$ after adaptation & $3.1 \times 10^{-5}$ -- $1.8 \times 10^{-1}$ \\
 & affine share of the induced output change & 1.4\% -- 99.5\% \\
Submersion (Sec.~\ref{sec:manifold}) & random $64$-row block of $\partial \pred|_\Msk / \partial\theta$ at full rank & 150/150 \\
 & $\sigma_{\min}/\sigma_{\max}$ of that block & $1.1 \times 10^{-4}$ -- $8.6 \times 10^{-3}$ \\
\bottomrule
\end{tabular}
}
\end{table}

\paragraph{The implemented objective.}
The implemented loss matches \eqref{eq:closed} to within $1.7\times10^{-6}$ relative deviation,
the detached and exact gradients have cosine at least $0.99998$, and both flat manifolds of
(C2) hold to floating-point accuracy: the output-affine invariance to $1.1\times10^{-6}$ and the
factorization invariance to $5\times10^{-3}$ relative deviation over the full group
$\mathrm{GL}(r)$---diagonal, permutation, orthogonal and general $M$ with condition numbers up to
$10^2$, drawn independently per adapted layer---not only over scalar rescalings
(Tab.~\ref{tab:analysis}).

\begin{figure}[!htbp]
\centering
\includegraphics[width=\linewidth]{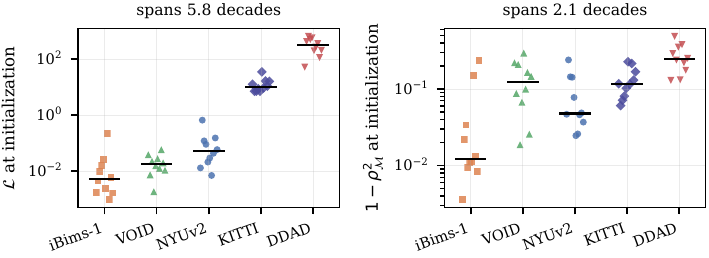}
\caption{\textbf{The objective at initialization, per sample and benchmark.} Bars mark medians.
\emph{Left}: the objective. \emph{Right}: the same 50 samples divided by $\Var_\Msk(y)$,
i.e., the dimensionless factor $1-\rho_\Msk^2$ of \eqref{eq:closed}.}
\label{fig:scale}
\end{figure}

\paragraph{(C1) The data scale.}
Fig.~\ref{fig:scale} shows the objective at initialization before and after dividing out
$\Var_\Msk(y)$: the span shrinks from about six orders of magnitude to about two. The remaining
spread is real---indoor scenes are closer to the frozen prediction than driving scenes---but
small enough that one dimensionless amplitude serves every benchmark (App.~\ref{sec:appx-sws}).

\begin{figure}[!htbp]
\centering
\includegraphics[width=\linewidth]{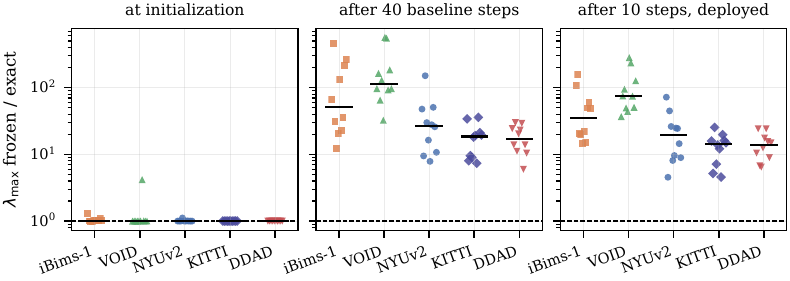}
\caption{\textbf{Largest Hessian eigenvalue with the scale and shift held constant, divided by
the same quantity with the alignment differentiated.} Bars mark medians over 50 images,
at the initialization, after the baseline's $40$ steps and after the $10$ warm-started steps
we deploy.}
\label{fig:aligncurv}
\end{figure}

\paragraph{(C2) Curvature must be computed with the alignment differentiated.}
Fig.~\ref{fig:aligncurv} measures this at three points on the trajectory. The two estimates
agree at the initialization, where the adapter's effect is exactly zero; after the baseline's
$40$ steps and after our $10$ warm-started steps they diverge alike, by a median factor of
$20$ (range 5--282) on the deployed $10$-step run. A frozen alignment reports curvature along
directions in which \eqref{eq:closed} is exactly flat, so any curvature-based diagnosis of this
objective, including of our own runs, must differentiate the alignment. All curvature
quantities here use Hessian-vector products with the alignment differentiated, extremal
eigenvalues by Lanczos iteration and traces by Hutchinson probes, restricted to the parameters
that can influence the depth output; in this backbone some adapter tensors feed only the
confidence head and would otherwise dilute every spectral quantity.

\begin{figure}[!htbp]
\centering
\includegraphics[width=\linewidth]{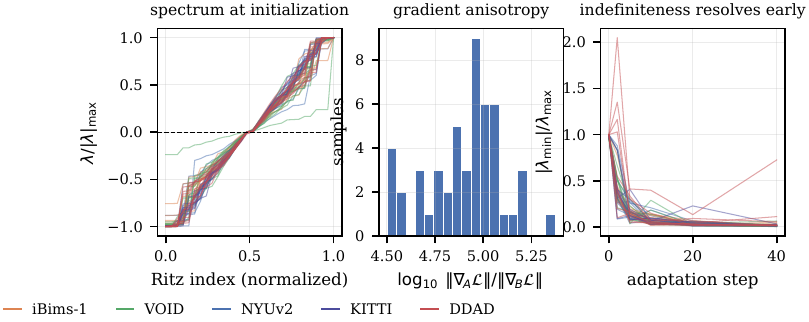}
\caption{\textbf{The Hessian at the zero initialization.} Computed over the 50-image
diagnostic subset. \emph{Left}: Ritz values, normalized by the largest magnitude, one curve per sample.
\emph{Middle}: the ratio of the two factors' gradient norms a millionth of the way along the
trajectory (at the initialization itself the denominator is exactly zero). \emph{Right}:
$|\lambda_{\min}|/\lambda_{\max}$ along the run.}
\label{fig:saddle}
\end{figure}

\paragraph{(C3) The degenerate initialization.}
Fig.~\ref{fig:saddle} shows (C3) in three views. The Ritz spectrum at the initialization is
close to antisymmetric on every sample (between $47\%$ and $50\%$ of the Ritz values are
negative), as expected when the cross block dominates the Hessian; the gradient-norm ratio
between the two factors is four to five orders of magnitude a millionth of the way along the
trajectory; and the indefiniteness resolves within a few steps, steps that a short budget can
ill afford. Tab.~\ref{tab:seed} shows the same effect through the seed: the spread over five
adapter initializations grows with the depth range, because the seed selects the escape
direction from the degenerate point. \sws{} overwrites both factors, so its spread is exactly
zero.

\begin{table}[!htbp]
\centering
\caption{\textbf{Standard deviation of MAE (m) over five adapter seeds.} Rows with \sws{} are
exactly zero because the warm start overwrites both factors.}
\label{tab:seed}
\footnotesize
\fittable{\begin{tabular}{lccccc}
\toprule
Method & iBims-1 & VOID & NYUv2 & KITTI & DDAD \\
\midrule
\lora{} & 0.0010 & 0.0032 & 0.0013 & 0.0198 & 0.1073 \\
\sws{} (either read-out) & 0.0000 & 0.0000 & 0.0000 & 0.0000 & 0.0000 \\
\bottomrule
\end{tabular}
}
\end{table}

\paragraph{The submersion hypothesis.}
The codimension of Prop.~\ref{prop:codim} is $m-2$ only where $\theta \mapsto \pred|_\Msk$ has
full row rank. The Jacobian is $m \times D$ with $D = 438{,}784$ and $m$ up to $18{,}966$, so neither
forming it nor computing its smallest singular value is affordable per sample, but a random
$64$-row block is. Full rank of every such block is necessary for the hypothesis but not
sufficient: a rank drop would falsify it, and full rank at healthy conditioning is evidence for
it. We measure at three points on the trajectory---the frozen model, the $10$-step state we
deploy and the $40$-step baseline endpoint---and every block is full rank ($150/150$
sample--point pairs), with $\sigma_{\min}/\sigma_{\max}$ between $1.1 \times 10^{-4}$ and
$8.6 \times 10^{-3}$: no sampled block shows a rank drop.

\begin{figure}[!htbp]
\centering
\includegraphics[width=\linewidth]{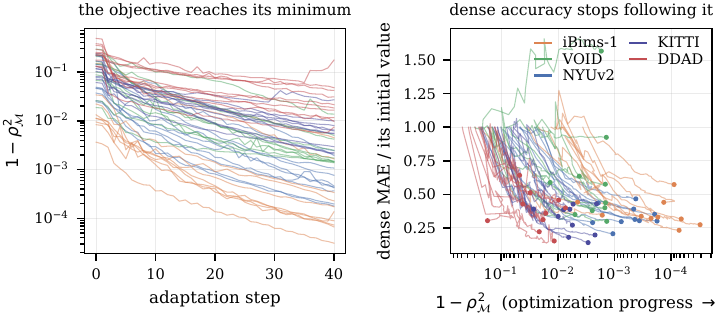}
\caption{\textbf{The objective along the run.} One curve per sample over the 50-image
diagnostic subset. \emph{Left}: the objective against the step count. \emph{Right}: dense
accuracy against the objective, so progress runs from left to right.}
\label{fig:saturate}
\end{figure}

\paragraph{Drift along the minimizer set.}
Fig.~\ref{fig:saturate} is the measurement behind Sec.~\ref{sec:manifold}. The objective
saturates early, yet dense accuracy continues to improve: the run keeps moving along
$\Theta^\star$, and the drift changes the unmeasured pixels. This is why the adaptation budget
matters even though the objective is essentially solved, and why choosing the point directly,
as \srp{} does, is much cheaper than buying the drift with more steps.

\FloatBarrier
\section{Ablations of the Propagation}
\label{sec:appx-srp}

\paragraph{One design decision at a time.}
Tab.~\ref{tab:solve} removes one decision from \srp{} at a time. Two rows put a constant back
and two do not. Writing the measurements as a penalty costs $18\%$ on the frozen
prediction even with $\mu$ tuned on the calibration set; replacing \eqref{eq:rank} by
$\exp(-\lVert \Delta I\rVert/\sigma)$ with a tuned $\sigma$ costs $12\%$; and both
reintroduce a per-benchmark constant in data units, which (C1) says does not transfer. A uniform
affinity costs $27\%$, more than any other single decision, and dropping the parametric
term costs $13\%$ of RMSE on DDAD. After $40$ steps all variants are close to full
\srp{}: these decisions carry information a long way, and after adaptation the residual is
small and local.

\begin{table}[!htbp]
\centering
\caption{\textbf{One design decision of \srp{} removed at a time.} Each variant is evaluated with
no adaptation and after $40$ steps. ``Const.''\ counts the values each variant must choose per benchmark; each variant
keeps the rest of \srp{} fixed. 895 images, stratified across the five benchmarks.}
\label{tab:solve}
\footnotesize
\fittable{\begin{tabular}{llcccccccccc}
\toprule
 & & \multicolumn{2}{c}{iBims-1} & \multicolumn{2}{c}{VOID} & \multicolumn{2}{c}{NYUv2} & \multicolumn{2}{c}{KITTI} & \multicolumn{2}{c}{DDAD} \\
\cmidrule(lr){3-4}\cmidrule(lr){5-6}\cmidrule(lr){7-8}\cmidrule(lr){9-10}\cmidrule(lr){11-12}
Variant & Const. & MAE & RMSE & MAE & RMSE & MAE & RMSE & MAE & RMSE & MAE & RMSE \\
\midrule
\multicolumn{12}{l}{\emph{no adaptation}} \\
\textbf{\srp{}} \emph{(ours)} & 0 & 0.040 & 0.162 & 0.193 & 0.693 & 0.056 & 0.143 & 0.339 & 1.435 & 1.484 & 6.404 \\
measurements as a penalty, tuned $\mu$ & 1 & 0.049 & 0.171 & 0.208 & 0.717 & 0.064 & 0.156 & 0.355 & 1.504 & 1.961 & 7.221 \\
affinity by bandwidth, tuned $\sigma$ & 1 & 0.046 & 0.167 & 0.198 & 0.698 & 0.061 & 0.148 & 0.367 & 1.520 & 1.666 & 6.567 \\
uniform affinity (no image, no depth) & 0 & 0.055 & 0.174 & 0.201 & 0.702 & 0.076 & 0.166 & 0.377 & 1.488 & 1.795 & 6.591 \\
without the parametric term \eqref{eq:trend} & 0 & 0.040 & 0.163 & 0.190 & 0.691 & 0.055 & 0.143 & 0.342 & 1.485 & 1.572 & 7.392 \\
\midrule
\multicolumn{12}{l}{\emph{after 40 adaptation steps}} \\
\textbf{\srp{}} \emph{(ours)} & 0 & 0.037 & 0.146 & 0.170 & 0.658 & 0.051 & 0.123 & 0.316 & 1.312 & 1.396 & 5.608 \\
measurements as a penalty, tuned $\mu$ & 1 & 0.038 & 0.146 & 0.171 & 0.659 & 0.051 & 0.123 & 0.316 & 1.311 & 1.420 & 5.620 \\
affinity by bandwidth, tuned $\sigma$ & 1 & 0.038 & 0.146 & 0.170 & 0.658 & 0.051 & 0.123 & 0.318 & 1.317 & 1.426 & 5.626 \\
uniform affinity (no image, no depth) & 0 & 0.038 & 0.146 & 0.170 & 0.658 & 0.052 & 0.124 & 0.316 & 1.305 & 1.440 & 5.599 \\
without the parametric term \eqref{eq:trend} & 0 & 0.037 & 0.146 & 0.170 & 0.659 & 0.051 & 0.123 & 0.316 & 1.313 & 1.396 & 5.608 \\
\bottomrule
\end{tabular}
}
\end{table}

\paragraph{Image structure and density.}
The uniform-affinity row asks what guided filtering~\citep{he2010guidedfilter} asks: what does
image structure buy? Sorting samples by measurement density sharpens the answer. On the same
runs as Tab.~\ref{tab:solve}, removing the image still costs $+3.1\%$ of MAE on the three
sparsest quartiles after adaptation, and nothing ($-0.1\%$) on the densest. Image
structure is what lets a residual travel, so it matters where measurements are sparse and the
residual is large. We do not gate the affinity on density, since the gate would be one more
constant to choose.

\paragraph{The solver budget is monotone.}
Tab.~\ref{tab:solver} solves \eqref{eq:harm} four ways: plain conjugate gradients at $400$ and
$3{,}200$ iterations, the cascadic cycle at the deployed fixed budget, and the same cycle stopped
on a residual tolerance. They bracket the same field, and the deployed budget is within
$1.1\%$ of convergence at a fraction of the cost. There is no interior optimum to find,
so the budget is a cost choice; and only the fixed-budget cycle has a loop length set by the
resolution alone, so only it can be captured as a graph.

\begin{table}[!htbp]
\centering
\caption{\textbf{One problem, four solvers.} Every row solves \eqref{eq:harm} on the frozen
prediction with the same affinity; the last row adds \eqref{eq:trend} to the deployed solve,
which is \srp{}. 6{,}479 images.}
\label{tab:solver}
\footnotesize
\fittable{\begin{tabular}{llcccccccccc}
\toprule
 & & \multicolumn{2}{c}{iBims-1} & \multicolumn{2}{c}{VOID} & \multicolumn{2}{c}{NYUv2} & \multicolumn{2}{c}{KITTI} & \multicolumn{2}{c}{DDAD} \\
\cmidrule(lr){3-4}\cmidrule(lr){5-6}\cmidrule(lr){7-8}\cmidrule(lr){9-10}\cmidrule(lr){11-12}
Solver & Iters & MAE & RMSE & MAE & RMSE & MAE & RMSE & MAE & RMSE & MAE & RMSE \\
\midrule
plain CG, $400$ iterations & $400$ & 0.040 & 0.162 & 0.148 & 0.533 & 0.054 & 0.142 & 0.343 & 1.498 & 1.661 & 8.040 \\
plain CG, $3{,}200$ iterations & $3{,}200$ & 0.039 & 0.162 & 0.147 & 0.530 & 0.054 & 0.142 & 0.342 & 1.482 & 1.514 & 6.724 \\
cascadic, residual tolerance $10^{-4}$ & adaptive & 0.039 & 0.162 & 0.147 & 0.531 & 0.054 & 0.142 & 0.342 & 1.483 & 1.514 & 6.726 \\
\textbf{cascadic, $4\times120$ (deployed)} & $480$ & 0.040 & 0.163 & 0.149 & 0.534 & 0.054 & 0.142 & 0.343 & 1.491 & 1.568 & 7.271 \\
\quad the same, $+$ \eqref{eq:trend} \emph{(\srp{})} & $480$ & 0.040 & 0.162 & 0.152 & 0.539 & 0.055 & 0.142 & 0.340 & 1.448 & 1.480 & 6.325 \\
\bottomrule
\end{tabular}
}
\end{table}

\paragraph{Reach.}
Tab.~\ref{tab:reach} is the measurement behind Sec.~\ref{sec:srp} and is reported per benchmark
because pooling destroys it. The maximum principle predicts not that the solve is worse than
adaptation but that the margin grows with distance, and it does so on exactly two benchmarks.
On NYUv2 the ratio falls slightly with distance, from $1.31$ to $1.14$; on iBims-1 it rises
mildly, from $0.97$ to $1.30$; and on VOID it stays near parity throughout, even though VOID has
the largest share of distant pixels of any benchmark (over a quarter beyond $16$ pixels):
distance alone does not make the ceiling bind. On DDAD the solve is better than parity within
$4$ pixels and $5.7\times$ worse between $16$ and $64$ (5.2--6.2);
\eqref{eq:trend} cuts that group to $2.3\times$ (2.1--2.5), a factor of
$2.5$ with disjoint intervals, and improves the nearer groups as well. KITTI shows the
same shape at less than half the size ($2.4\times$ in the same group before
\eqref{eq:trend}), as its measurement pattern predicts: $96\%$ of its evaluated
pixels lie within $4$ pixels of a measurement. A better graph or a longer solve could not
supply this, since the converged field is still bounded by \eqref{eq:maxprin}; the parametric
term is not.

\begin{table}[!htbp]
\centering
\caption{\textbf{Error against distance to the nearest measurement.} Per benchmark, the first
row is the share of evaluated pixels in each group; the others give the summed squared error of
\srp{} on the frozen prediction, with and without \eqref{eq:trend}, divided by that of a
$40$-step \lora{} run on the same group (bold: above $2$). Groups with fewer than 10{,}000
evaluated pixels print the count instead. 2{,}767 images, 95--768 per benchmark.}
\label{tab:reach}
\footnotesize
\fittable{\begin{tabular}{llcccc}
\toprule
 & & \multicolumn{4}{c}{distance to the nearest measurement (px)} \\
\cmidrule(lr){3-6}
Benchmark & & $0$--$4$ & $4$--$16$ & $16$--$64$ & $>64$ \\
\midrule
\multirow{3}{*}{iBims-1} & evaluated pixels & 14.2\% & 78.1\% & 7.6\% & \emph{892\,px} \\
 & \srp{} without \eqref{eq:trend} & 0.97 & 1.14 & 1.30 & --- \\
 & \srp{} & 0.96 & 1.13 & 1.27 & --- \\
\midrule
\multirow{3}{*}{VOID} & evaluated pixels & 11.6\% & 60.5\% & 25.4\% & 2.6\% \\
 & \srp{} without \eqref{eq:trend} & 1.06 & 1.02 & 1.04 & 1.07 \\
 & \srp{} & 1.06 & 1.01 & 1.04 & 1.09 \\
\midrule
\multirow{3}{*}{NYUv2} & evaluated pixels & 27.5\% & 71.9\% & 0.6\% & 0 \\
 & \srp{} without \eqref{eq:trend} & 1.31 & 1.26 & 1.14 & --- \\
 & \srp{} & 1.31 & 1.25 & 1.14 & --- \\
\midrule
\multirow{3}{*}{KITTI} & evaluated pixels & 96.0\% & 4.0\% & 0.037\% & \emph{1\,px} \\
 & \srp{} without \eqref{eq:trend} & 1.13 & 1.24 & \textbf{2.36} & --- \\
 & \srp{} & 1.07 & 1.15 & 1.97 & --- \\
\midrule
\multirow{3}{*}{DDAD} & evaluated pixels & 39.8\% & 58.5\% & 1.7\% & \emph{672\,px} \\
 & \srp{} without \eqref{eq:trend} & 0.95 & 1.46 & \textbf{5.67} & --- \\
 & \srp{} & 0.86 & 1.19 & \textbf{2.30} & --- \\
\bottomrule
\end{tabular}
}
\end{table}

\paragraph{Where the prior is placed.} The solve \eqref{eq:harm} applies a smoothness prior to the correction rather than to the depth.
Tab.~\ref{tab:tv} tests the assumption behind this directly, with no accuracy metric in the
loop, by measuring how strongly each field concentrates its total variation at the ground-truth
depth discontinuities; the correction field scored is the one \srp{} produces on the deployed
configuration. Indoors, ground-truth depth concentrates its variation at its discontinuities
far more than any computed field ($18.5$--$65.4$ against at most $6.6$), so a prior that is smooth
away from edges is a poor description of depth. Among the three computed fields, the correction
is the least concentrated on four of the five benchmarks and over the pooled set ($3.31$ against
$4.56$ and $5.43$), and on DDAD it is level with the frozen prediction ($3.99$ against $3.87$): it is
the field for which the prior fits best, which is why \eqref{eq:harm} is applied to it. The
separation is narrowest on the driving benchmarks, whose ground truth is itself least
concentrated at its discontinuities ($5.0$--$5.6$).

\begin{table}[!htbp]
\centering
\caption{\textbf{Concentration $C(u)$ of a field's total variation at ground-truth
discontinuities.} The mean of $\lVert\nabla u\rVert_1$ over the $5\%$ of valid pixels with the
largest ground-truth gradient, divided by the mean over the rest. \emph{All} pools every sample.
50 images.}
\label{tab:tv}
\footnotesize
\setlength{\tabcolsep}{4pt}
\fittable{\begin{tabular}{lcccccc}
\toprule
Field & iBims-1 & VOID & NYUv2 & KITTI & DDAD & All \\
\midrule
Ground-truth depth & 55.74 & 65.36 & 18.54 & 5.57 & 5.04 & 14.60 \\
Frozen prediction & 6.62 & 3.17 & 5.47 & 4.95 & 3.87 & 4.56 \\
Adapted prediction $\hat f$ & 6.11 & 3.35 & 6.28 & 5.61 & 5.50 & 5.43 \\
Correction field $\delta^\star$ & 3.79 & 2.09 & 2.78 & 3.89 & 3.99 & 3.31 \\
\bottomrule
\end{tabular}
}
\end{table}

\paragraph{How many parametric terms.}
Tab.~\ref{tab:basis} sweeps nested bases on the frozen prediction, scoring each by dense error
and by the distance decomposition of Tab.~\ref{tab:reach}, so that reach and generalization are
separate columns of one run. More terms reach further: the $16$--$64$ group on DDAD falls from
$2.09\times$ \lora{}'s squared error at three terms to $1.86\times$ at four,
and no further ($1.87\times$ at six to fourteen terms). More terms also generalize worse:
dense error rises monotonically from $93.4\%$ of \lora{}'s at three terms to
$95.8\%$ at fourteen, because every added coefficient is fitted on a measured set that
is a biased sample of the scene. The balance tips between three and four terms; we keep three,
since the four-term gain is $0.2$ in a group holding under $2\%$ of DDAD's pixels, against
a dense cost paid on all of them. A trend in \emph{position}, as in regression kriging, gives
$94.0\%$ dense error and $6.16\times$ in the $16$--$64$ group: a trend on the grid is
another field on the grid and says nothing about a far pixel that the solve does not. An
IRLS-reweighted fit is worse on both columns, as expected when residuals are structured rather
than contaminated. Re-estimating the affinity from the corrected depth and solving again changes
nothing---the second solve reproduces the first to seven digits---because the first solve
already matches every measurement exactly, so the second receives a residual that is zero on
$\Msk$: a property of the constrained form that a penalized solve would not have.

\begin{table}[!htbp]
\centering
\caption{\textbf{Size of the parametric basis.} Measured on the frozen prediction with everything
else in \srp{} fixed. ``MAE\,/\,\lora{}'' is the mean over the five benchmarks of dense error relative
to a $40$-step \lora{} run; the last column is the squared-error ratio in DDAD's $16$--$64$\,px
group (Tab.~\ref{tab:reach}). $x, y$ are pixel coordinates on $[-1,1]$. 495 images, up to 100 per
benchmark.}
\label{tab:basis}
\footnotesize
\fittable{\begin{tabular}{lccc}
\toprule
Parametric term & Terms & MAE\,/\,\lora{} & $16$--$64$\,px\,/\,\lora{} \\
\midrule
\textbf{$[1, \hat f^2, \log \hat f]$ \eqref{eq:trend}, deployed} & 3 & 93.4\% & 2.09 \\
\quad $+\ \hat f$ & 4 & 93.9\% & 1.86 \\
\quad $+\ x, y$ & 6 & 94.2\% & 1.87 \\
\quad $+\ x^2, y^2, xy$ & 9 & 94.9\% & 1.87 \\
\quad $+\ \hat f\,\{x, y, x^2, y^2, xy\}$ & 14 & 95.8\% & 1.87 \\
$[1, x, y]$, a trend in \emph{position} & 3 & 94.0\% & 6.16 \\
\quad $+\ x^2, y^2, xy$ & 6 & 94.4\% & 6.18 \\
three terms, IRLS-reweighted fit & 3 & 93.6\% & 4.58 \\
three terms, affinity re-estimated and re-solved & 3 & 93.4\% & 2.09 \\
\bottomrule
\end{tabular}
}
\end{table}

\FloatBarrier
\section{Ablations of the Warm Start}
\label{sec:appx-sws}

\paragraph{Components.}
Tab.~\ref{tab:sws-ablation} separates \sws{} into its parts at both budgets. The LoRA-GA-style
row is the controlled contrast---same gradient, factorization, optimizer and budget, with the
step canceled---and aligning the initialization this way is worth $3.9\%$ at $40$
steps and $8.6\%$ at $10$. Taking the step, scaled per layer and with its computed
length, adds $3.4\%$ and $5.3\%$, for totals of $7.3\%$ and
$13.9\%$. Cutting the budget to a quarter roughly doubles the value of the alignment:
a long run finds a good subspace by itself and a short one does not. A random nonzero
up-projection of the same magnitude is neutral (Tab.~\ref{tab:interventions}), so what matters is
where the starting point is, not that it is nonzero.

\begin{table}[!htbp]
\centering
\caption{\textbf{Components of the warm start.} The indented row adds to the one above. Columns: median
final MAE relative to \lora{}, mean seed standard deviation and the fraction of samples ending worse
than the frozen backbone (over the core). The LoRA-GA-style row cancels the step. 247
images, stratified across the five benchmarks.}
\label{tab:sws-ablation}
\footnotesize
\fittable{\begin{tabular}{lcccccc}
\toprule
 & \multicolumn{3}{c}{40 steps} & \multicolumn{3}{c}{10 steps} \\
\cmidrule(lr){2-4}\cmidrule(lr){5-7}
Variant & MAE\,/\,\lora{} & seed sd & fail & MAE\,/\,\lora{} & seed sd & fail \\
\midrule
\lora{} (no warm start) & 1.000 & 0.0225 & 0.0\% & 1.000 & 0.0488 & 0.5\% \\
Spectral step, no per-layer scale & 0.974 & 0.0000 & 0.0\% & 0.913 & 0.0000 & 1.5\% \\
\quad $+$ per-layer scale $\Fnorm{W_l}$ (\sws{}) & 0.927 & 0.0000 & 0.0\% & 0.861 & 0.0000 & 1.5\% \\
LoRA-GA-style init~\citep{wang2024loraga} (step canceled) & 0.961 & 0.0000 & 0.0\% & 0.914 & 0.0000 & 1.5\% \\
\bottomrule
\end{tabular}
}
\end{table}

\paragraph{The per-layer scale.}
Dropping $\Fnorm{W_l}$ from \eqref{eq:sws} costs $5.2\%$ at $10$ steps and
$4.7\%$ at $40$. With the amplitude computed by \eqref{eq:tau}, the absolute variant
no longer diverges, since \eqref{eq:tau} scales its coefficient to whatever ray it is given, so
the per-layer factor now buys accuracy rather than stability: it sets where the step lands. At
$10$ steps, every warm-started variant ends worse than the frozen backbone on $1.5\%$
of samples against the baseline's $0.5\%$; at $40$ steps no variant does. Every
\sws{} variant has a seed standard deviation of exactly zero, against $0.049$ for the
baseline at $10$ steps.

\begin{figure}[!htbp]
\centering
\includegraphics[width=\linewidth]{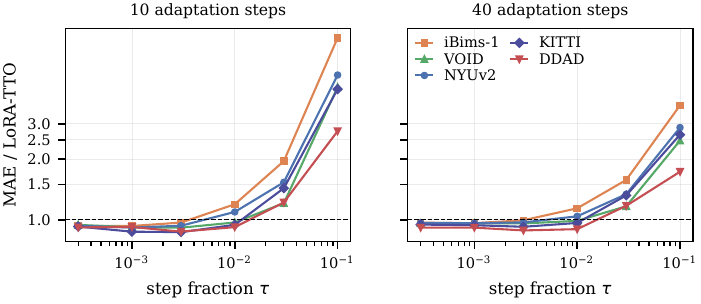}
\caption{\textbf{Median MAE relative to \lora{} at the same budget, against a fixed amplitude, the step fraction
$\tau$ in \eqref{eq:sws}.} One $\tau$ is shared by every sample, with no excursion cap. \emph{Left}: $10$ steps.
\emph{Right}: $40$ steps. 250 images.}
\label{fig:tau}
\end{figure}

\paragraph{Why one $\tau$ can be shared.}
Fig.~\ref{fig:tau} sweeps a fixed amplitude. The curve is flat below a threshold and a cliff
above it, and the collapse begins at the same $\tau$ on benchmarks whose depth scales differ by
two orders of magnitude, which is what the per-layer $\Fnorm{W_l}$ provides. The curve is also
asymmetric: the first-order model of \eqref{eq:tau} errs low, since the tangent at the origin
underestimates the ray's maximizer, and low is the inexpensive side. At $10$ steps the choice
within the flat region is worth a few points; at $40$ it is worth a fraction of one, because a
longer run absorbs the initialization's error.

\paragraph{The amplitude rule and the refinement count.}
Tab.~\ref{tab:amplitude} compares \eqref{eq:tau} with greedy search and with one fixed $\tau$, on
495 images drawn evenly from the five evaluation splits; no separate validation set was
used, and App.~\ref{sec:appx-heldout} repeats the comparison with each benchmark held out
(Tab.~\ref{tab:lobo}) and checks the final configuration on the 4{,}907 images outside every
development subset (Tab.~\ref{tab:heldout}). The comparison gives the search every advantage. The grid we started from was
too low---its upper half is selected on 1 of 495 samples, and
$48\%$ of samples select its lowest point---so we also report a grid placed on the
useful range and scored on held-out pixels or on the training loss. The best grid and a single
fixed value land within $0.7$ points of each other, with no setting reliably best,
and \eqref{eq:tau} with two refinements is better than the best search by $0.3$ points
at $10$ steps and $0.6$ at $40$, for $4$ forward passes against $6$
and with nothing to choose. The only column in which a grid is ahead, by $0.3$ points, is the
$10$-step run without the read-out, which no reported configuration uses. Two refinements beat
none, one and three, so the count has an interior optimum; unlike a constant in data units it is one
dimensionless integer, fixed for all five benchmarks at once. A third re-solve lands closer to
the ray's $\rho^2_\Msk$ maximizer yet is worse on dense error, because that maximizer is defined
on the measured pixels; this is also why the count is fixed globally and never selected per
sample.

\begin{table}[!htbp]
\centering
\caption{\textbf{How the amplitude is set.} Same direction \eqref{eq:sws} and recipe throughout;
``Const.''\ counts the values to choose and ``Fwd.''\ the forward passes spent. Entries are the
mean over the five benchmarks of MAE relative to \lora{} at the same budget ($100$ is parity).
495 images.}
\label{tab:amplitude}
\footnotesize
\fittable{\begin{tabular}{lccccc}
\toprule
 & & & \multicolumn{2}{c}{10 steps} & 40 steps \\
\cmidrule(lr){4-5}\cmidrule(lr){6-6}
How the amplitude is set & Const. & Fwd. & alone & $+$ \srp{} & $+$ \srp{} \\
\midrule
greedy search, grid on the useful range & 1 & 6 & 89.2 & 96.8 & 98.2 \\
\quad the wider grid we started from & 1 & 7 & 88.7 & 96.8 & 98.5 \\
one fixed $\tau$, shared by every sample & 1 & 1 & 89.3 & 97.4 & 98.8 \\
\eqref{eq:tau}, no refinement & 0 & 1 & 92.5 & 96.9 & 98.0 \\
\quad $+$ 1 refinement & 0 & 3 & 89.5 & 96.6 & 98.0 \\
\quad\textbf{$+$ 2 refinements} \emph{(deployed)} & 0 & 4 & 89.0 & 96.5 & 97.7 \\
\quad $+$ 3 refinements & 0 & 5 & 89.3 & 96.7 & 97.9 \\
\bottomrule
\end{tabular}
}
\end{table}

\FloatBarrier
\section{Design Alternatives That Do Not Help}
\label{sec:appx-alt}

Sec.~\ref{sec:manifold} predicts that interventions improving the \emph{optimization} should not
help, because the optimization already succeeds, and neither should interventions regularizing
in \emph{parameter} space, because the dimension of $\Theta^\star$ is set by the number of
unmeasured pixels. App.~\ref{sec:appx-alt-predicted} tests both classes;
App.~\ref{sec:appx-alt-other} reports alternatives the analysis does not rule out but that a
reader may consider. All use the backbone and budget of the main results, on the subsets stated in the captions.

\subsection{Alternatives the analysis predicts should not help}
\label{sec:appx-alt-predicted}

\paragraph{Optimization and parameter-space regularization.}
Tab.~\ref{tab:interventions} lists twelve interventions applied to \lora{}. Ten are at or above
parity: a warm-up with a cosine schedule, factor-asymmetric learning
rates~\citep{hayou2024loraplus} (the more aggressive ratio fails on every sample),
sharpness-aware minimization~\citep{foret2021sam}, parameter averaging, early stopping on
held-out measured pixels, a proximal term, two output trust regions and a nonzero random
up-projection. The two below parity move along the exactly flat directions of (C2) and are within
half a percent, the expected outcome for changes the objective cannot see. \lora{} is also at
its step-size ceiling (Tab.~\ref{tab:optim}): one grid step higher diverges, and raising the
learning rate \emph{after} the warm start diverges at the same threshold, so the ceiling is a
property of the objective and not of the starting point. The same table bounds the effect of
weight decay at about $2\%$; it is the one setting in which our runs could have differed
from \lora{}, and they do not.

\begin{table}[!htbp]
\centering
\caption{\textbf{Optimization and parameter-space interventions.} Each is applied to \lora{} with
everything else fixed. Columns: median final MAE relative to \lora{} and the fraction of runs ending worse
than the frozen backbone. 50 images, over the core.}
\label{tab:interventions}
\footnotesize
\fittable{\begin{tabular}{lcc}
\toprule
Intervention & MAE\,/\,\lora{} & fail \\
\midrule
Warm-up $+$ cosine schedule & 1.079 & 0.0\% \\
LoRA$+$, ratio 4~\citep{hayou2024loraplus} & 1.316 & 13.3\% \\
LoRA$+$, ratio 16~\citep{hayou2024loraplus} & 15.127 & 100.0\% \\
Sharpness-aware minimization~\citep{foret2021sam} & 1.002 & 0.0\% \\
Parameter EMA over the last half of the run & 1.017 & 0.0\% \\
Early stopping on held-out measured pixels & 1.086 & 0.8\% \\
Proximal term toward the initialization & 1.206 & 0.0\% \\
Output trust region, $0.10$ & 1.009 & 0.0\% \\
Output trust region, $0.20$ & 1.000 & 0.0\% \\
Nonzero random up-projection & 1.008 & 0.0\% \\
Balanced factor norms & 0.996 & 0.0\% \\
Affine-free adapter effect & 0.997 & 0.0\% \\
\bottomrule
\end{tabular}
}
\end{table}

\paragraph{The outer loss.}
The closed form holds because the inner alignment is the \emph{exact} minimizer of a squared
error. Tab.~\ref{tab:loss} replaces the loss with four alternatives, and each is worse: $\ell_1$,
Huber, $\ell_1$+RMSE and SILog each leave at least one benchmark worse than no adaptation at all,
and both robust losses fail on both driving benchmarks, where the measured set
contains the returns a robust loss is designed to discount. None of them admits the closed form
\eqref{eq:closed}, so none of the analysis applies to them.

\begin{table}[!htbp]
\centering
\begin{minipage}[t]{0.52\linewidth}
\centering
\caption{\textbf{Outer loss in \eqref{eq:obj}.} Median final MAE relative to the frozen
backbone ($>1$ is worse than no adaptation). 50 images.}
\label{tab:loss}
\footnotesize
\fittable{\begin{tabular}{lccccc}
\toprule
Loss & iBims-1 & VOID & NYUv2 & KITTI & DDAD \\
\midrule
MSE (default) & 0.335 & 0.505 & 0.309 & 0.330 & 0.421 \\
$\ell_1$+RMSE & 0.321 & 2.304 & 0.339 & 0.378 & 0.819 \\
Huber & 0.327 & 0.513 & 0.315 & 1.991 & 3.162 \\
$\ell_1$ & 0.484 & 3.852 & 0.535 & 2.509 & 3.387 \\
SILog & 0.570 & 1.691 & 1.355 & 6.267 & 3.552 \\
\bottomrule
\end{tabular}
}
\end{minipage}\hfill
\begin{minipage}[t]{0.45\linewidth}
\centering
\caption{\textbf{Optimizer and learning rate of the inherited recipe.} Median final MAE
relative to the frozen backbone, over the core. 50 images.}
\label{tab:optim}
\footnotesize
\fittable{\begin{tabular}{lc}
\toprule
Optimizer, learning rate & MAE / frozen backbone \\
\midrule
SGD, $10^{-1}$ & 0.999 \\
AdamW, $10^{-3}$ & 0.402 \\
AdamW, $3\!\times\!10^{-3}$ & 0.329 \\
AdamW, $10^{-2}$ \emph{(reference)} & 0.306 \\
Adam (no weight decay), $10^{-2}$ & 0.312 \\
AdamW, $3\!\times\!10^{-2}$ & 3.838 \\
\bottomrule
\end{tabular}
}
\end{minipage}
\end{table}

\paragraph{Restricting the adapter to a subspace.}
The trajectory looks low-dimensional---its first principal component carries most of the
variance, and the merged weight update has stable rank near one---which suggests freezing the
down-projection at the leading singular vectors of $G_l$ and training only the up-projection.
This reaches $1.15\times$ \lora{}'s MAE (median over the benchmarks). The subspace a run
needs is not fixed: the initial subspace captures a median $0.41$ of the final update's
energy. (We do not quote principal angles: in an ambient space this large, the largest angle
between two rank-8 subspaces is near $90^\circ$ for almost any pair.) The optimizer rotates its
own subspace during the run, which is why \sws{} uses the spectral step only as an
initialization and then frees both factors.

\paragraph{Extrapolating the trajectory.}
If the path from the initialization to the endpoint were straight, moving further along it would
keep improving. At $\theta_0 + 1.5(\theta_T - \theta_0)$, however, the objective is $79\times$
its value at $\theta_T$ (median over the benchmarks): the path along $\Theta^\star$ is curved, and its
endpoint cannot be reached early by extrapolation.

\subsection{Other alternatives}
\label{sec:appx-alt-other}

\paragraph{Folding the solve into the adapter.}
If the field the solve produces is the one we want, one could train the adapter to produce it
and drop the second stage. Tab.~\ref{tab:folding} reports five training losses in three families,
under one protocol. \emph{Distillation} replaces the objective by agreement with the converged
field, in the closed form of \eqref{eq:closed} so that a constant prediction is the loss's
maximum rather than a collapse it can fall into; a variant adds the measurements back with a
relative weight. \emph{Graph regularization} keeps the published objective and adds the Dirichlet
energy of \eqref{eq:harm} as a penalty. \emph{The constrained functional} replaces the objective
by the one \eqref{eq:harm} minimizes, with and without the published loss alongside. No training
loss yields an adapter that can stand in for the solve: every one is beaten by applying the solve
to the frozen prediction and training nothing ($1.123$), and the best adapter any of them
produces reaches $1.146$ against \sws{}'s $1.088$. With the solve retained, the differences
shrink to a few percent and none reaches \sws{}. In every family the sweep selects the end of the
grid at which the added term is weakest. The reason is representational: asked only to copy the
converged field, with no measurements to satisfy, the adapter reaches $1.737$ while the field
itself is at $1.123$. A rank-8 perturbation of the backbone's weights cannot express what a
free field on the pixel grid can, so the two stages stay separate.

\begin{table}[!htbp]
\centering
\caption{\textbf{Training the adapter to replace the solve.} Median MAE relative to
\sws{}~$+$~\srp{}, without and with the solve at read-out; each row is the best of its own sweep,
and ``seed sd'' is over three seeds. 25 images, 5 per benchmark (20 from the
calibration set and 5 from DDAD's evaluation split).}
\label{tab:folding}
\footnotesize
\fittable{\begin{tabular}{lccc}
\toprule
 & \multicolumn{2}{c}{adapter alone} & $+$ \srp{} \\
\cmidrule(lr){2-3}\cmidrule(lr){4-4}
What the adapter is asked to fit & MAE & seed sd & MAE \\
\midrule
\emph{no adapter} & 3.201 & 0.000 & 1.123 \\
the published objective~\citep{seo2026efficient} & 1.174 & 0.026 & 1.031 \\
\quad the converged field, copied & 1.737 & 0.016 & 1.097 \\
\quad that field and the measurements & 1.146 & 0.030 & 1.029 \\
\quad the measurements and a graph penalty & 1.199 & 0.019 & 1.027 \\
\quad the constrained functional & 1.356 & 0.029 & 1.008 \\
\quad that and the published objective & 1.196 & 0.031 & 1.011 \\
\textbf{\sws{}} \emph{(ours)} & 1.088 & 0.000 & 1.000 \\
\bottomrule
\end{tabular}
}
\end{table}

\paragraph{A second-order energy.}
The intermediate distance is where the solve trails adaptation (App.~\ref{sec:appx-srp}), and
\eqref{eq:maxprin} says why. A textbook alternative to a parametric term is the second-order
energy $\lVert \Lap\,\delta \rVert^2$, whose null space contains linear functions, so its
minimizer extrapolates a trend instead of relaxing to a weighted average---the thin plate to
\eqref{eq:harm}'s membrane. It has no constant either. At this budget, however, it does not
converge: squaring the Laplacian squares its condition number, and after $1{,}200$ conjugate-gradient
iterations, an order of magnitude more work than the deployed solve, the field more than
$64$ pixels from any measurement is still exactly the frozen prediction. Its error is worse
than \eqref{eq:harm}'s on every benchmark and improves monotonically with iterations, the
signature of an unconverged solver rather than a wrong prior. A multigrid preconditioner for the
biharmonic operator might change this; at this budget, the three-unknown regression of
\eqref{eq:trend} is the form of extrapolation that converges.

\paragraph{Solving in log space.}
Solving \eqref{eq:harm} on $\log \hat f$ makes the correction multiplicative. It helps on DDAD and
costs more elsewhere, for a net loss, and it is redundant once \eqref{eq:trend} is present: the
$\log \hat f$ term already carries the global part of a multiplicative correction.

\paragraph{Confidence-modulated regularization.}
Weighting the data term of the penalized form (Sec.~\ref{sec:srp}) by the backbone's confidence
output selects zero modulation on every sample. The question does not arise for \eqref{eq:harm},
which has no data term to weight.

\begin{figure}[!htbp]
\centering
\includegraphics[width=\linewidth]{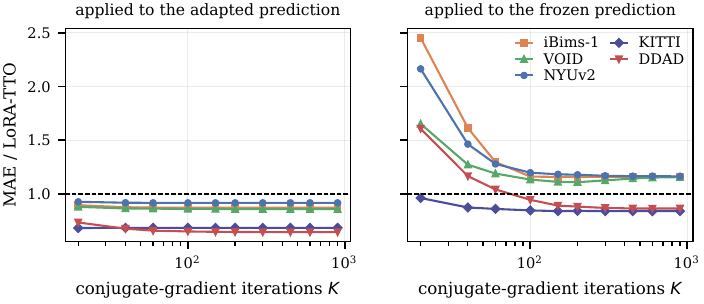}
\caption{\textbf{Iteration count $K$ of the \emph{penalized} solve of Sec.~\ref{sec:srp}, per
benchmark.} Median MAE relative to \lora{} against the conjugate-gradient iteration count, on
an adapted and on the frozen prediction. 50 images.}
\label{fig:kcurve}
\end{figure}

\paragraph{The iteration count of the penalized form.}
In the penalized form the iteration count is a regularizer: truncating conjugate gradients
restricts the correction to a Krylov subspace in which the well-conditioned components appear
first~\citep{hansen1998rankdeficient}, so accuracy rises and then falls and some $K$ has to be
chosen (Tab.~\ref{tab:k}, Fig.~\ref{fig:kcurve}). The optimum differs between benchmarks and
between the adapted and frozen predictions, where the residual is larger and less local.
Choosing $K$ \emph{per sample} from held-out measured pixels is worse than one fixed count,
because the measured pixels share the bias the solve exists to correct: held-out sparse
agreement is not an unbiased estimate of dense error. This argument concerns
\emph{estimating} dense error from measured pixels; \eqref{eq:tau} estimates nothing but solves a
stationarity condition of the training objective, which is defined on the measured pixels by
construction, as are all the gradients that follow it. None of this applies to \eqref{eq:harm},
whose accuracy is monotone in the work spent (Tab.~\ref{tab:solver}): its budget can be set from
the latency available.

\begin{table}[!htbp]
\centering
\caption{\textbf{Iterations $K$ of the \emph{penalized} solve.} Median MAE relative to \lora{},
over the core. \best{Bold and underlined} marks the value chosen on the calibration set.
50 images.}
\label{tab:k}
\footnotesize
\setlength{\tabcolsep}{4pt}
\fittable{\begin{tabular}{lccccccc}
\toprule
CG iterations $K$ & 20 & 40 & 60 & 100 & 200 & 450 & 900 \\
\midrule
applied to an adapted prediction & 0.817 & \best{0.774} & 0.774 & 0.776 & 0.777 & 0.777 & 0.776 \\
applied to the frozen prediction & 1.862 & 1.291 & 1.154 & 1.082 & \best{0.999} & 0.970 & 0.955 \\
\bottomrule
\end{tabular}
}
\end{table}

\FloatBarrier
\section{Qualitative Results}
\label{sec:appx-qual}

\begin{figure}[!htbp]
\centering
\includegraphics[width=0.9\linewidth]{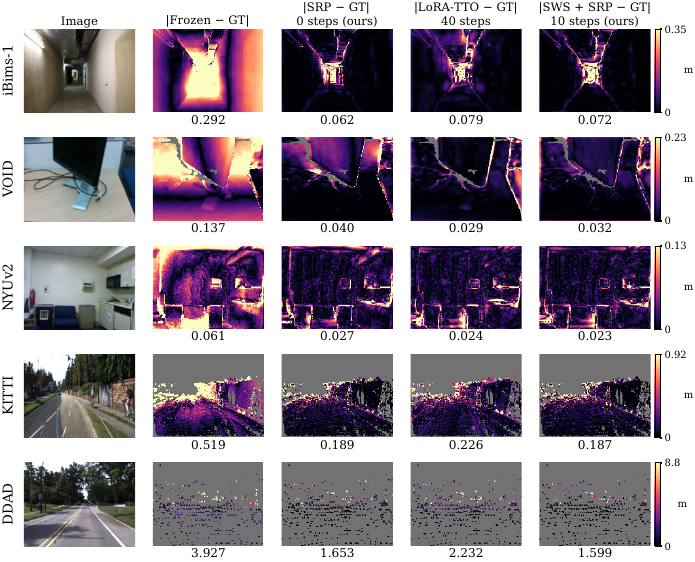}
\caption{\textbf{Absolute error (m) on the samples of Fig.~\ref{fig:qualitative}.} The scale
(\texttt{magma}) is shared within a row and saturates at the $95$th percentile of its four maps;
error is defined only where ground truth exists (pixels without ground truth are gray), which on KITTI and DDAD
is $19\%$ and $1.0\%$ of pixels.}
\label{fig:qualitative-err}
\end{figure}

\begin{figure}[!htbp]
\centering
\includegraphics[width=0.9\linewidth]{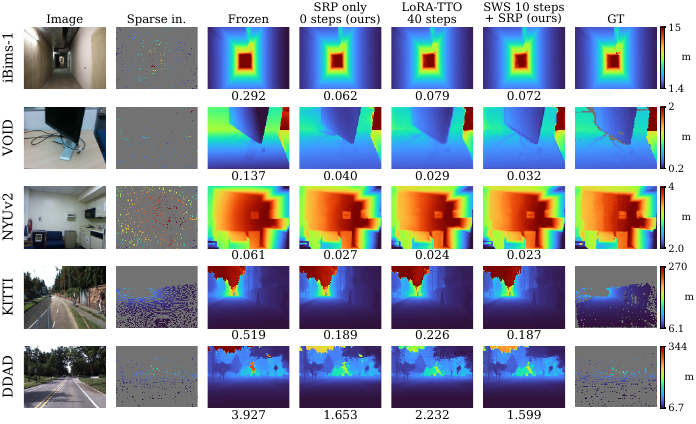}
\caption{\textbf{Predicted depth on one sample per benchmark.} Four configurations, with the
per-sample MAE (m) under each panel. The color scale (\texttt{turbo}) is fixed per
row and shown in meters at the right; masked panels are dilated for visibility.}
\label{fig:qualitative}
\end{figure}

Fig.~\ref{fig:qualitative-err} shows what four configurations leave behind on one sample per
benchmark, and Fig.~\ref{fig:qualitative} what they predict. The frozen backbone's error is
regional rather than edge-shaped: whole surfaces sit at the wrong distance even after the
scale-and-shift fit, which is what \eqref{eq:closed} says survives the alignment. After
adaptation or the read-out, the error no longer fills regions; it collects along depth
discontinuities and, on the driving rows, at the pixels farthest from any measurement, the group
Tab.~\ref{tab:reach} isolates and \eqref{eq:trend} exists to reach. The read-out with no backward
pass already produces a residual of the second kind, a visual counterpart of the claim that
choosing among the minimizers is a different job from descending toward them. The quantitative
evidence is Tab.~\ref{tab:main} and Tab.~\ref{tab:reach}.

\FloatBarrier
\section{Limitations}
\label{sec:appx-limits}

\paragraph{Intermediate distance.}
Without adaptation, \srp{} trails the $40$-step run between $4$ and $16$ pixels from a
measurement on every benchmark (Tab.~\ref{tab:reach}), and this group holds most evaluated pixels
on four of the five. A smooth interpolant cannot invent the structure missing there, which is
what the $10$ adaptation steps supply; an extrapolating field that converges at this budget
(App.~\ref{sec:appx-alt-other}) is the natural next step.

\paragraph{Feature-driven sparsity.}
VOID is the benchmark our full-budget configurations move least, the only one where the
parametric term costs rather than helps, and the one where a richer basis does most harm. One
property accounts for all three: its points come from a visual-inertial tracker and sit on
trackable corners and textured structures, so the measured set concentrates where the affinity
graph of \eqref{eq:harm} is weakly connected and is a biased sample of the scene, and anything
fitted or selected on it generalizes least well. Affinities that stay informative under such
patterns are the clearest open problem this work leaves.

\paragraph{The tail and the operating point.}
Without adaptation, \srp{} improves MAE on three of the five benchmarks but is behind on RMSE on
all five, because a squared metric is priced by the tail the field reaches least well; the
$10$-step configuration is the first point on the curve that is ahead on both. At $10$ steps,
\sws{} ends worse than the frozen backbone on $1.5\%$ of samples against the
baseline's $0.5\%$ (App.~\ref{sec:appx-sws}).

\paragraph{Scope.}
The pipeline is not real-time: at the adaptation-free point it runs at 9.2 frames per
second at VGA and 1.0 at FHD on one RTX 4090, and we state cost claims as ratios against
the baseline. The analysis covers any aligned masked regression, but we instantiate it with
low-rank adaptation on two backbones, and of the diffusion-prior baselines only
Marigold-DC is run (App.~\ref{sec:appx-setup}).

\section{Additional Comparisons}
\label{sec:appx-more}

\subsection{\texorpdfstring{Marigold-DC on a fixed subset}{Marigold-DC on a fixed subset}}
\label{sec:appx-mdc-subset}

The best published numbers of Marigold-DC average a 10-sample ensemble, which costs ten times
one sample: $219$~s per VGA image under the protocol of Tab.~\ref{tab:cost}, against $22.0$~s. We run the ensemble, and one sample at four further seeds, on a subset fixed before
any result was seen: all 95 iBims-1 images and 40 per other benchmark, drawn by a fixed seed
(255 images). Settings and scoring are those of Tab.~\ref{tab:main} (App.~\ref{sec:appx-setup}).
One sample varies little across seeds, by at most $2.7\%$ of the mean on any benchmark
(Tab.~\ref{tab:mdc-subset}). The ensemble narrows the gap to the baseline from $1.33$ to $1.07$
of its MAE over the five benchmarks (from $1.23$ to $1.06$ of its RMSE) and leads every
configuration on NYUv2's MAE, tying the $40$-step configuration on its RMSE; on the other four
benchmarks it trails the $40$-step baseline and both adapted configurations of ours on both
metrics. On the same images, our $10$-step configuration reaches $0.87$ of the baseline's MAE.

\begin{table}[!htbp]
\centering
\caption{\textbf{Marigold-DC on a fixed subset.} MAE and RMSE in meters on 255 images
(all 95 of iBims-1, 40 per other benchmark), fixed before any result was seen. One sample is the
mean over five seeds, with its standard deviation below it; the ensemble averages 10 samples.
$^\dagger$Denoising steps per sample. The other rows are those of Tab.~\ref{tab:main} on the same
images.}
\label{tab:mdc-subset}
\footnotesize
\fittable{\begin{tabular}{lccccccccccc}
\toprule
 & & \multicolumn{2}{c}{iBims-1} & \multicolumn{2}{c}{VOID} & \multicolumn{2}{c}{NYUv2} & \multicolumn{2}{c}{KITTI} & \multicolumn{2}{c}{DDAD} \\
\cmidrule(lr){3-4}\cmidrule(lr){5-6}\cmidrule(lr){7-8}\cmidrule(lr){9-10}\cmidrule(lr){11-12}
Method & Steps & MAE & RMSE & MAE & RMSE & MAE & RMSE & MAE & RMSE & MAE & RMSE \\
\midrule
Marigold-DC, 1 sample & 50$^\dagger$ & 0.063 & 0.202 & 0.069 & 0.365 & 0.058 & 0.145 & 0.552 & 1.550 & 2.876 & 7.534 \\
\quad seed standard deviation &  & 0.000 & 0.001 & 0.002 & 0.004 & 0.001 & 0.002 & 0.010 & 0.022 & 0.036 & 0.025 \\
Marigold-DC, 10-sample ensemble & 50$^\dagger$ & 0.045 & 0.166 & 0.063 & 0.354 & 0.049 & 0.127 & 0.423 & 1.349 & 2.257 & 6.101 \\
\midrule
\lora{}~\citep{seo2026efficient} & 40 & 0.041 & 0.146 & 0.056 & 0.336 & 0.055 & 0.129 & 0.414 & 1.260 & 1.877 & 5.671 \\
\srp{} \emph{(ours)} & 0 & 0.040 & 0.162 & 0.069 & 0.364 & 0.057 & 0.149 & 0.322 & 1.341 & 1.398 & 6.034 \\
\sws{} $+$ \srp{} \emph{(ours)} & 10 & 0.041 & 0.144 & 0.052 & 0.328 & 0.055 & 0.131 & 0.305 & 1.203 & 1.337 & 5.361 \\
\sws{} $+$ \srp{} \emph{(ours)} & 40 & 0.037 & 0.146 & 0.049 & 0.327 & 0.051 & 0.127 & 0.292 & 1.214 & 1.290 & 5.499 \\
\bottomrule
\end{tabular}
}
\end{table}

\subsection{\texorpdfstring{Density, structured missingness and placement}{Density, structured missingness and placement}}
\label{sec:appx-density}

The sweep was fixed before any result was seen: 100 NYUv2 and 100 KITTI images evenly spaced
through the images outside every development subset (Tab.~\ref{tab:decisions}); random
thinning to 50, 20, 10, 5 and 2\% of the protocol's points, nested so that each level is a
subset of the denser one; KITTI's lidar thinned to 32, 16, 8, 4, 2 and 1 lines by pitch
binning, as in the channel study of \citet{seo2026efficient}; and four structured patterns,
removing the left half of the image, a central box covering a quarter of it, or the farthest
25\% or 50\% of the measurements. Every configuration is re-run on the thinned measurements;
the dense ground truth is unchanged, and at full density every value reproduces
Tab.~\ref{tab:main} per image. \lora{} uses one seed here.

Tab.~\ref{tab:density} reports every pattern. The $10$-step configuration leads the baseline's
MAE on KITTI under all of them, by $18\%$ with a single scan line and by $19\%$ and $22\%$ when
the farthest quarter or half of the points is removed, consistent with the reach the parametric
term adds (Tab.~\ref{tab:reach}). On NYUv2 it stays within $2\%$ at every density and leads at the sparsest ($-4.0\%$ at
10 points), and the $40$-step configuration leads at every density. Among the structured
patterns, the one in which the $10$-step configuration trails by more than $1\%$ is NYUv2 with
its left half removed ($+3.8\%$): a single room's 475 points then leave half of the image with no
measurement to constrain it, and the $40$-step configuration closes most of the gap ($+1.1\%$).
\srp{} alone, the adaptation-free point, keeps its MAE lead on KITTI down to $10\%$ of the points or
16 lines and falls behind at sparser settings except a single line; on NYUv2 it is level at full
density and falls behind as points thin ($+12\%$ at 10 points), and it trails under every
structured pattern. This is the regime the $10$ adaptation steps cover.
At equal budget, adding \srp{} to \lora{} never costs more than $0.2\%$ of MAE ($0.4\%$ of RMSE)
in any pattern.

\textbf{Placement.} VOID's points sit on corners and textured structure (App.~\ref{sec:appx-limits}). To isolate placement on data with dense ground truth, we place the measurements of the same 100 NYUv2 images only on image corners---Shi--Tomasi corners of the grey image at the evaluation resolution (quality level $0.001$, minimum distance 3 pixels), strongest first, skipping pixels without ground truth---at each image's protocol count and at VOID's median share of the image, $0.23\%$ or 154 points, against random placement at the same counts (Tab.~\ref{tab:density}). Moving the points onto corners changes the $10$-step configuration's MAE ratio by at most one point and lowers its RMSE ratio, and the $40$-step configuration leads on both metrics under every placement.

\begin{table}[!htbp]
\centering
\caption{\textbf{Density, structured missingness and placement.} \lora{} (40 steps, one seed) in
meters, and MAE / RMSE of each configuration relative to it on the same images and thinned
measurements; below 1 is better. ``Points'' is the median number of measurements kept.
100 held-out images per benchmark. Placement rows put the measurements only on image corners, at each image's protocol count or at 154 points (VOID's median share of the image), or at random at 154 points; random placement at the protocol's count is the NYUv2 100\% row.}
\label{tab:density}
\footnotesize
\fittable{\begin{tabular}{llrcccccc}
\toprule
 & & & \multicolumn{2}{c}{\lora{} (40)} & \multicolumn{4}{c}{MAE / RMSE relative to \lora{} (40)} \\
\cmidrule(lr){4-5}\cmidrule(lr){6-9}
Setting & Pattern & Points & MAE & RMSE & \srp{} (0) & \sws{}$+$\srp{} (10) & \sws{}$+$\srp{} (40) & \lora{}$+$\srp{} (40) \\
\midrule
NYUv2{,} random thinning & 100\% & 475 & 0.056 & 0.131 & 0.992 / 1.128 & 0.972 / 0.978 & 0.896 / 0.969 & 0.924 / 0.983 \\
 & 50\% & 238 & 0.069 & 0.156 & 1.001 / 1.071 & 0.987 / 0.978 & 0.927 / 0.965 & 0.967 / 0.993 \\
 & 20\% & 95 & 0.090 & 0.192 & 1.007 / 1.025 & 1.017 / 1.003 & 0.970 / 0.990 & 0.992 / 0.999 \\
 & 10\% & 48 & 0.108 & 0.212 & 1.030 / 1.042 & 0.996 / 0.999 & 0.964 / 0.984 & 0.997 / 0.999 \\
 & 5\% & 24 & 0.133 & 0.251 & 1.040 / 1.024 & 0.977 / 0.986 & 0.956 / 0.970 & 0.998 / 1.000 \\
 & 2\% & 10 & 0.180 & 0.303 & 1.119 / 1.080 & 0.960 / 0.971 & 0.982 / 0.988 & 1.002 / 1.001 \\
\midrule
KITTI{,} random thinning & 100\% & 16{,}642 & 0.444 & 1.371 & 0.761 / 1.052 & 0.720 / 0.944 & 0.697 / 0.964 & 0.715 / 0.965 \\
 & 50\% & 8{,}322 & 0.445 & 1.409 & 0.839 / 1.090 & 0.784 / 0.951 & 0.755 / 0.971 & 0.774 / 0.971 \\
 & 20\% & 3{,}329 & 0.472 & 1.516 & 0.923 / 1.138 & 0.855 / 0.968 & 0.814 / 0.978 & 0.838 / 0.983 \\
 & 10\% & 1{,}665 & 0.520 & 1.653 & 0.951 / 1.142 & 0.880 / 0.966 & 0.831 / 0.962 & 0.868 / 0.988 \\
 & 5\% & 832 & 0.561 & 1.789 & 1.007 / 1.160 & 0.928 / 0.981 & 0.877 / 0.968 & 0.917 / 0.992 \\
 & 2\% & 334 & 0.649 & 2.067 & 1.073 / 1.163 & 0.963 / 0.986 & 0.921 / 0.972 & 0.960 / 0.996 \\
\midrule
KITTI{,} scan lines & 32 lines & 8{,}454 & 0.467 & 1.487 & 0.872 / 1.104 & 0.815 / 0.960 & 0.790 / 0.981 & 0.814 / 0.983 \\
 & 16 lines & 4{,}098 & 0.536 & 1.740 & 0.953 / 1.105 & 0.893 / 0.972 & 0.864 / 0.982 & 0.894 / 0.994 \\
 & 8 lines & 2{,}003 & 0.652 & 2.037 & 1.002 / 1.142 & 0.902 / 0.957 & 0.881 / 0.973 & 0.923 / 0.988 \\
 & 4 lines & 1{,}097 & 0.774 & 2.271 & 1.058 / 1.118 & 0.961 / 0.970 & 0.929 / 0.964 & 0.979 / 0.999 \\
 & 2 lines & 510 & 1.020 & 2.538 & 1.166 / 1.152 & 0.996 / 0.972 & 0.957 / 0.962 & 0.987 / 0.999 \\
 & 1 line & 377 & 1.556 & 3.776 & 0.792 / 0.848 & 0.820 / 0.847 & 0.993 / 0.967 & 1.001 / 1.001 \\
\midrule
NYUv2{,} structured & left half removed & 238 & 0.112 & 0.226 & 1.140 / 1.097 & 1.038 / 1.013 & 1.011 / 1.014 & 0.989 / 0.998 \\
 & central box removed & 358 & 0.071 & 0.159 & 1.029 / 1.096 & 0.997 / 0.989 & 0.940 / 0.984 & 0.961 / 0.993 \\
 & farthest 25\% removed & 356 & 0.127 & 0.286 & 1.233 / 1.173 & 1.008 / 1.002 & 1.002 / 1.012 & 0.985 / 0.998 \\
 & farthest 50\% removed & 238 & 0.234 & 0.432 & 1.342 / 1.229 & 0.942 / 0.940 & 0.957 / 0.963 & 0.997 / 1.000 \\
\midrule
KITTI{,} structured & left half removed & 8{,}622 & 0.740 & 2.185 & 1.155 / 1.190 & 0.962 / 1.004 & 0.929 / 1.008 & 0.937 / 0.995 \\
 & central box removed & 10{,}404 & 0.856 & 2.582 & 1.149 / 1.147 & 0.945 / 0.959 & 0.924 / 0.964 & 0.947 / 0.997 \\
 & farthest 25\% removed & 12{,}484 & 2.518 & 7.032 & 1.217 / 1.133 & 0.814 / 0.820 & 0.862 / 0.880 & 0.998 / 1.004 \\
 & farthest 50\% removed & 8{,}325 & 3.841 & 8.728 & 1.105 / 1.007 & 0.780 / 0.779 & 0.853 / 0.861 & 1.001 / 1.001 \\
\midrule
NYUv2{,} placement & corners & 475 & 0.086 & 0.166 & 1.078 / 1.069 & 0.982 / 0.959 & 0.925 / 0.966 & 0.961 / 0.988 \\
 & random{,} 154 points & 154 & 0.077 & 0.170 & 1.017 / 1.060 & 1.008 / 0.990 & 0.948 / 0.973 & 0.984 / 0.998 \\
 & corners{,} 154 points & 154 & 0.134 & 0.234 & 1.073 / 1.031 & 1.015 / 0.973 & 0.943 / 0.952 & 0.994 / 0.999 \\
\bottomrule
\end{tabular}
}
\end{table}

\subsection{\texorpdfstring{The fast bilateral solver on the residual}{The fast bilateral solver on the residual}}
\label{sec:appx-fbs}

We apply the fast bilateral solver (FBS)~\citep{barron2016fbs} where \srp{} acts: to the
residual on the measured pixels, with confidence one there and zero elsewhere, adding its output
to the aligned prediction. We use the authors' reference implementation, with its
conjugate-gradient budget raised from 25 to 200 iterations. FBS has three constants: the spatial
and color bandwidths of its bilateral grid, and the weight $\lambda$ of its smoothness term
against the data penalty. We give it every advantage, an oracle that no deployed method has: for
each benchmark and each prediction, all three are chosen on the dense ground truth of 30 images from 455 combinations---$\sigma_s$
from 1 to 16 pixels and $\sigma_c$ from 4 to 256 intensity levels in powers of two, and 13 values
of $\lambda$ from $10^{-4}$ to $256$---and evaluated on the 895 images of Tab.~\ref{tab:solve},
which include the tuning images. Both bandwidths are searched to their ends: $\sigma_s = 1$ puts
one pixel in each grid cell, and $\sigma_c = 256$ leaves a single intensity bin, removing the
color axis.

On the frozen prediction, tuned FBS trails \srp{} on four benchmarks, by $12$--$24\%$ of MAE
(Tab.~\ref{tab:fbs}). Averaged over the five, it reaches $1.03$ of the MAE of $40$-step \lora{},
against $0.93$ for \srp{}: with no adaptation step, \srp{} leads the $40$-step baseline and tuned
FBS does not. With the reference parameters, FBS reaches $3.14$. It leads \srp{} on VOID by $1.7\%$, where our parametric term itself costs
$1.2\%$ (Tab.~\ref{tab:solve}) and FBS has none. After $40$ steps the residual is small and
local, and, as for the variants of Tab.~\ref{tab:solve}, every tuned read-out is close: FBS is
within $2.1\%$ of \srp{} on every benchmark.

\begin{figure}[!htbp]
\centering
\includegraphics[width=\linewidth]{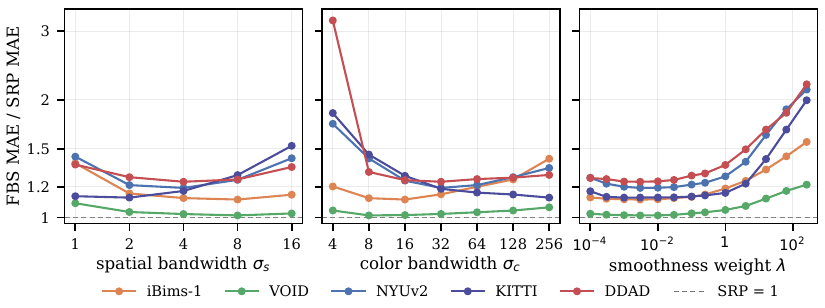}
\caption{\textbf{Tuning the fast bilateral solver,} frozen prediction. Mean MAE of FBS
relative to \srp{} on the 30 tuning images per benchmark, against each constant, every point
the best over the other two; log scale, and $1$ is \srp{}.}
\label{fig:fbs-tuning}
\end{figure}

Fig.~\ref{fig:fbs-tuning} shows how FBS reaches its best. The chosen $\lambda$ lies inside
the range searched, between $0.001$ and $0.01$, on the flat floor of every curve, where the data
penalty acts almost as the constraint that \eqref{eq:harm} imposes exactly; larger values smooth
across the measurements. On KITTI the ground truth selects $\sigma_c = 256$: FBS discards its color
axis and becomes a purely spatial smoother. Each of the three constants is chosen per benchmark,
and on ground truth; \srp{} chooses none (C1).

\begin{table}[!htbp]
\centering
\caption{\textbf{The fast bilateral solver on the residual.} MAE and RMSE in meters on the
895 images of Tab.~\ref{tab:solve}, with no adaptation and after $40$ steps. ``Const.''\ counts
the values chosen per benchmark, as in Tab.~\ref{tab:solve}; for tuned FBS they are chosen on the
dense ground truth of 30 of these images per benchmark and prediction. The reference parameters
are those of the authors' code.}
\label{tab:fbs}
\footnotesize
\fittable{\begin{tabular}{llcccccccccc}
\toprule
 & & \multicolumn{2}{c}{iBims-1} & \multicolumn{2}{c}{VOID} & \multicolumn{2}{c}{NYUv2} & \multicolumn{2}{c}{KITTI} & \multicolumn{2}{c}{DDAD} \\
\cmidrule(lr){3-4}\cmidrule(lr){5-6}\cmidrule(lr){7-8}\cmidrule(lr){9-10}\cmidrule(lr){11-12}
Read-out & Const. & MAE & RMSE & MAE & RMSE & MAE & RMSE & MAE & RMSE & MAE & RMSE \\
\midrule
\multicolumn{12}{l}{\emph{no adaptation}} \\
\textbf{\srp{}} \emph{(ours)} & 0 & 0.040 & 0.162 & 0.193 & 0.693 & 0.056 & 0.143 & 0.339 & 1.435 & 1.484 & 6.404 \\
FBS, tuned per benchmark on ground truth & 3 & 0.045 & 0.170 & 0.189 & 0.689 & 0.064 & 0.154 & 0.381 & 1.594 & 1.841 & 8.402 \\
FBS, reference parameters & 0 & 0.127 & 0.254 & 0.242 & 0.747 & 0.134 & 0.236 & 1.836 & 3.622 & 8.893 & 16.430 \\
\midrule
\multicolumn{12}{l}{\emph{after 40 adaptation steps}} \\
\textbf{\srp{}} \emph{(ours)} & 0 & 0.037 & 0.146 & 0.170 & 0.658 & 0.051 & 0.123 & 0.316 & 1.312 & 1.396 & 5.608 \\
FBS, tuned per benchmark on ground truth & 3 & 0.038 & 0.146 & 0.170 & 0.658 & 0.051 & 0.123 & 0.316 & 1.305 & 1.426 & 5.648 \\
FBS, reference parameters & 0 & 0.045 & 0.149 & 0.175 & 0.661 & 0.055 & 0.125 & 0.560 & 1.549 & 2.554 & 6.580 \\
\bottomrule
\end{tabular}
}
\end{table}

\end{document}